\pdfoutput=1

\documentclass[11pt]{article}

\usepackage{latex/acl}

\usepackage{times}
\usepackage{latexsym}

\usepackage[T1]{fontenc}

\usepackage[utf8]{inputenc}

\usepackage{microtype}

\usepackage{inconsolata}

\usepackage{multirow}
\usepackage{subcaption}
\usepackage{graphicx}
\usepackage{amssymb}
\usepackage{amsthm}
\usepackage{amsmath,amssymb,amsfonts}
\usepackage{booktabs}
\usepackage{pifont}
\usepackage[ruled, linesnumbered, vlined]{algorithm2e}
\usepackage{verbatim}
\SetKwInput{KwInput}{Input}                
\SetKwInput{KwOutput}{Output}
\SetKwProg{Fn}{Procedure}{}{end}
\colorlet{mygreen}{gray!40!green}

\title{zrLLM: Zero-Shot Relational Learning on Temporal Knowledge Graphs with Large Language Models}

\author{Zifeng Ding\thanks{Equal contribution.}$^{1,2}$, Heling Cai\footnotemark[1]$^{1}$, Jingpei Wu$^{1}$, Yunpu Ma$^{1}$, \\ \textbf{Ruotong Liao$^{1,3}$, Bo Xiong\thanks{Corresponding author.}$^{4}$, Volker Tresp\footnotemark[2]$^{1}$} \\
$^{1}$LMU Munich
$^{2}$Siemens AG \\ $^{3}$Munich Center for Machine Learning (MCML) $^{4}$University of Stuttgart\\
\texttt{\{zifeng.ding, heling.cai\}@campus.lmu.de}\\
\texttt{\{jingpei.wu,liao,tresp\}@dbs.ifi.lmu.de,}\\
\texttt{cognitive.yunpu@gmail.com}, \texttt{bo.xiong@ki.uni-stuttgart.de}\\}

\begin{document}
\maketitle
\begin{abstract}
Modeling evolving knowledge over temporal knowledge graphs (TKGs) has become a heated topic. Various methods have been proposed to forecast links on TKGs. Most of them are embedding-based, where hidden representations are learned to represent knowledge graph (KG) entities and relations based on the observed graph contexts. Although these methods show strong performance on traditional TKG forecasting (TKGF) benchmarks, they face a strong challenge in modeling the unseen zero-shot relations that have no prior graph context. In this paper, we try to mitigate this problem as follows. We first input the text descriptions of KG relations into large language models (LLMs) for generating relation representations, and then introduce them into embedding-based TKGF methods. LLM-empowered representations can capture the semantic information in the relation descriptions. This makes the relations, whether seen or unseen, with similar semantic meanings stay close in the embedding space, enabling TKGF models to recognize zero-shot relations even without any observed graph context. Experimental results show that our approach helps TKGF models to achieve much better performance in forecasting the facts with previously unseen relations, while still maintaining their ability in link forecasting regarding seen relations.
\end{abstract}

\section{Introduction}
\label{sec: intro}

Knowledge graphs (KGs) represent world knowledge with a collection of facts in the form of $(s, r, o)$ triples, where in each fact, $s$, $o$ are the subject and object entities and $r$ is the relation between them. Temporal knowledge graphs (TKGs) are introduced by further specifying the time validity. Each TKG fact is denoted as a quadruple $(s,r,o,t)$, where $t$ (a timestamp or a time period) provides temporal constraints. Since world knowledge is ever-evolving, TKGs are more expressive in representing dynamic factual information and have drawn increasing interest in a wide range of downstream tasks, e.g., natural language question answering over TKGs \cite{DBLP:conf/acl/SaxenaCT20,DBLP:conf/semweb/DingLQWHMMCLHT23}.

In recent years, there has been an increasing number of works 
paying attention to forecasting future facts in TKGs, i.e., TKG forecasting (TKGF) or TKG extrapolated link prediction (LP). Most of them are embedding-based, where entity and relation representations are learned with the help of the observed graph contexts. Although traditional embedding-based TKGF methods show impressive performance on current benchmarks, they share a common limitation. In these works, models are trained on the TKG facts regarding a set of relations $\mathcal{R}$, and they are only expected to be evaluated on the facts containing the relations in $\mathcal{R}$. They cannot handle any zero-shot unseen relation $r \notin \mathcal{R}$ because no graph context regarding unseen relations exists in the training data and thus no reasonable relation representations can be learned. In the forecasting scenario, as time flows, new knowledge is constantly introduced into a TKG, making it expand in size. This increases the chance of encountering newly-emerged relations, and therefore, it is meaningful to improve embedding-based TKGF methods to be more adaptive to zero-shot relations.

With the increasing scale of pre-trained language models (LMs), LMs become large LMs (LLMs). Recent studies find that LLMs have shown emerging abilities in various aspects \cite{DBLP:journals/tmlr/WeiTBRZBYBZMCHVLDF22} and can be taken as strong semantic knowledge bases (KBs) \cite{DBLP:conf/emnlp/PetroniRRLBWM19}. Inspired by this, we try to enhance the performance of embedding-based TKGF models over zero-shot relations with an approach consisting of the following three steps: (1) Based on the relation text descriptions provided in TKG datasets, we first use an LLM 
to produce an enriched relation description (ERD) with more details for each KG relation (Sec. \ref{sec: enrich description}). (2) We then generate the relation representations by leveraging another LLM, i.e., T5-11B \cite{2020t5}. We input ERDs into T5's encoder and transform its output into relation representations of TKGF models (Sec. \ref{sec: alignment}). (3) We design a relation history learner (RHL) to capture historical relation patterns, where we leverage LLM-empowered relation representations
to better reason over zero-shot relations (Sec. \ref{sec: rhl}). With these steps, we align the natural language space provided by LLMs to the embedding space of TKGF models, rather than letting models learn relation representations solely from observed graph contexts. 
Even without any observed associated facts, 
zero-shot relations can be represented with LLM-empowered representations that contain semantic information. We term our approach as zrLLM since it is used to enhance \textbf{z}ero-shot \textbf{r}elational learning on TKGF models by using \textbf{LLM}s.

We experiment zrLLM on seven recent embedding-based TKGF models and evaluate them on three new datasets constructed specifically for studying TKGF regarding zero-shot relations. Our contribution is three-folded: (1) To the best of our knowledge, this is the first work trying to study zero-shot relational learning in TKGF. (2) We design an LLM-empowered approach zrLLM and manage to enhance various recent embedding-based TKGF models in reasoning over zero-shot relations. (3) Experimental results show that zrLLM helps to substantially improve all considered TKGF models' abilities in forecasting the facts containing unseen zero-shot relations, while still maintaining their ability in link forecasting regarding seen relations.

\section{Preliminaries}
\subsection{Related Work}
\paragraph{Traditional TKG Forecasting Methods.}
Traditional TKGF methods are trained to forecast the facts containing the KG relations (and entities) seen in the training data, regardless of the case where zero-shot relations (or entities) appear as new knowledge arrives. These methods can be categorized into two types: embedding-based and rule-based. Embedding-based methods learn hidden representations of KG relations and entities, and perform link forecasting based on them. Most existing embedding-based methods, e.g., \cite{DBLP:conf/emnlp/JinQJR20,DBLP:conf/emnlp/HanDMGT21,DBLP:conf/sigir/LiJLGGSWC21,DBLP:conf/ijcai/LiS022,DBLP:conf/icde/Liu0X0023}, learn evolutional entity and relation representations from the historical TKG information by jointly employing graph neural networks \cite{DBLP:conf/iclr/KipfW17} and recurrent neural structures, e.g., GRU \cite{DBLP:conf/emnlp/ChoMGBBSB14}. Some other approaches \cite{DBLP:conf/iclr/HanCMT21,sun-etal-2021-timetraveler,DBLP:conf/acl/LiJGLGWC20} start from each LP query\footnote{A TKG  LP query is denoted as $(s,r,?,t)$ (object prediction query) or $(?, r, o, t)$ (subject prediction query).} and traverse the temporal history in a TKG to search for the prediction answer. There also exist some methods, e.g., \cite{DBLP:conf/aaai/ZhuCFCZ21,DBLP:conf/aaai/XuO0F23}, that achieve forecasting based on the appearance of historical facts. 
Compared with embedding-based TKGF approaches, rule-based TKGF has still not been extensively explored. One popular rule-based TKGF method is TLogic \cite{DBLP:conf/aaai/LiuMHJT22}. It extracts temporal logic rules from TKGs and uses a symbolic reasoning module for LP. Based on it, ALRE-IR \cite{DBLP:conf/emnlp/MeiYCJ22} proposes an adaptive logical rule embedding model to encode temporal logical rules into rule representations. This makes ALRE-IR both a rule-based and an embedding-based method. Rule-based TKGF methods have strong ability in reasoning over zero-shot unseen entities connected by the seen relations, however, they are not able to handle unseen relations since the learned rules are strongly bounded by the observed relations. 

\paragraph{Inductive Learning on TKGs.}
Inductive learning on TKGs refers to developing models that can handle the relations and entities unseen in the training data. 
Most of TKG inductive learning methods are based on few-shot learning, e.g., \cite{ding2022few,DBLP:conf/nips/0007TYL19,DBLP:conf/pkdd/DingWLMT23,DBLP:conf/akbc/MirtaheriR0MG21,DBLP:conf/ijcnn/DingHWMHT23,DBLP:conf/ijcnn/DingHWMHT23,DBLP:journals/datamine/MaMMZL023}. They first compute inductive representations of newly-emerged entities or relations based on $K$-associated facts ($K$ is a small number, e.g., 1 or 3), and then use them to predict other facts regarding few-shot elements. One limitation of these works is that the inductive representations cannot be learned without the $K$-shot examples, making them hard to solve the zero-shot problems. Different from few-shot learning methods, SST-BERT \cite{DBLP:conf/sigir/ChenXS0D23} pre-trains a time-enhanced BERT \cite{DBLP:conf/naacl/DevlinCLT19} and proves its inductive power over unseen entities but has not shown its ability in reasoning zero-shot relations. Another recent work MTKGE \cite{DBLP:conf/www/ChenXS0D23} is able to concurrently deal with both unseen entities and relations. However, it requires a support graph containing a substantial number of data examples related to the unseen entities and relations, which is far from the zero-shot setting.

\paragraph{TKG Reasoning with Language Models.}
\label{sec: TKG with LLM}
Recently, more and more works have introduced LMs into TKG reasoning.
SST-BERT pre-trains an LM on a corpus of training TKGs for fact reasoning.
ECOLA \cite{han-etal-2023-ecola} aligns facts with additional fact-related texts and enhances TKG reasoning with BERT-encoded language representations. 
PPT \cite{DBLP:conf/acl/XuLPJP23} converts TKGF into the pre-trained LM masked token prediction task and finetunes a BERT for TKGF.
Apart from them, one recent work \cite{DBLP:journals/corr/abs-2305-10613} explores in-context learning (ICL) \cite{DBLP:conf/nips/BrownMRSKDNSSAA20} with LLMs to predict future facts without finetuning. 
Another recent work GenTKG \cite{DBLP:journals/corr/abs-2310-07793} finetunes Llama2-7B \cite{DBLP:journals/corr/abs-2302-13971}, and let it directly generate the LP answer in TKGF. 

Although previous works have shown success of LMs in TKG reasoning, they have limitations: (1) None of them has studied whether LMs, in particular LLMs, can be used to better reason zero-shot relations. (2) By only using ICL, LLMs are beaten by traditional TKGF methods in performance \cite{DBLP:journals/corr/abs-2305-10613}. The performance can be greatly improved by finetuning LLMs \cite{DBLP:journals/corr/abs-2310-07793}, but finetuning LLMs requires huge computational resources. (3) Since LMs are pre-trained with a huge corpus originating from diverse information sources, it is inevitable that they have already seen the world knowledge before they are used to solve TKG reasoning tasks. Most popular TKGF benchmarks are constructed with the facts before 2020 (ICEWS14/18/05-15 \cite{DBLP:conf/emnlp/JinQJR20}). The facts inside are based on the world knowledge before 2019, which means LMs might have encountered them in their training corpus, posing a threat of information leak to the LM-driven TKG reasoning models. To this end, we (1) draw attention to studying the impact of LLMs on zero-shot relational learning in TKGs; (2) make a compromise between performance and computational efficiency by not finetuning LMs or LLMs but adapting the LLM-provided semantic information to non-LM-based TKGF methods; (3) construct new benchmarks whose facts are all happening from 2021 to 2023, which avoids the threat of information leak when we utilize T5-11B that was released in 2020.

\subsection{Definitions and Task Formulation}
\noindent
\textbf{Definition 1 (TKG).} Let $\mathcal{E}$, $\mathcal{R}$, $\mathcal{T}$ denote a set of entities, relations and timestamps, respectively. A TKG $\mathcal{G} = \{(s,r,o,t)\} \subseteq \mathcal{E} \times \mathcal{R} \times \mathcal{E} \times \mathcal{T}$ is a set of temporal facts where each fact is represented with a fact quadruple $(s,r,o,t)$.

\noindent
\textbf{Definition 2 (TKG Forecasting).} Assume we have a ground truth TKG $\mathcal{G}_{gt}$ that contains all the true facts. Given an LP query $(s_q, r_q, ?, t_q)$ (or $(o_q, r_q,?, t_q)$), TKGF requires the models to predict the missing object $o_q$ (or subject $s_q$) based on the facts observed before the query timestamp $t_q$, i.e., $\mathcal{O} = \{(s,r,o,t_i) \in \mathcal{G}_{gt}|t_i< t_q\}$.

\noindent
\textbf{Definition 3 (Zero-Shot TKG Forecasting).} Assume we have a ground truth TKG $\mathcal{G}_{gt} \subseteq \mathcal{E} \times \mathcal{R} \times \mathcal{E} \times \mathcal{T}$, where $\mathcal{R}$ can be split into seen $\mathcal{R}_{se}$ and unseen $\mathcal{R}_{un}$ relations ($\mathcal{R} = \mathcal{R}_{se} \cup \mathcal{R}_{un}, \mathcal{R}_{se} \cap \mathcal{R}_{un} = \emptyset$). Given an LP query $(s_q, r_q, ?, t_q)$ (or $(o_q, r_q,?, t_q)$) whose query relation $r_q \in \mathcal{R}_{un}$, models are asked to predict the missing object $o_q$ (or subject $s_q$) based on the facts $\mathcal{O} = \{(s,r_i,o,t_i) \in \mathcal{G}_{gt}|t_i< t_q, r_i \in \mathcal{R}_{se}\}$ containing seen relations and happening before $t_q$.
\begin{figure*} 
    \centering
  \subfloat[Training pipeline of zrLLM-enhanced model.\label{fig: model_train}]{%
       \includegraphics[width=0.53\textwidth]{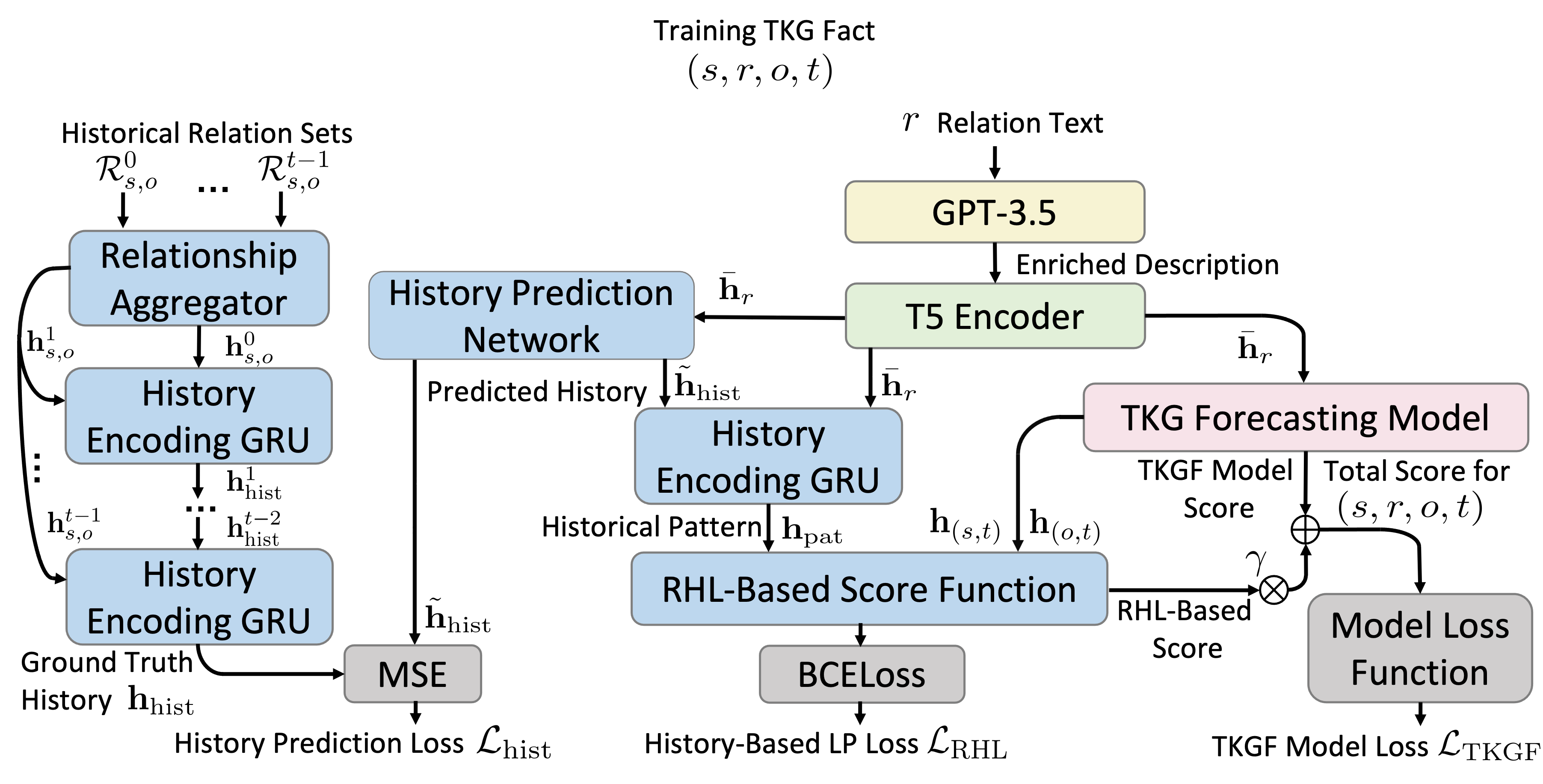}}
    \hfill
  \subfloat[Evaluation pipeline of zrLLM-enhanced model.\label{fig: model_eval}]{%
        \includegraphics[width=0.42\textwidth]{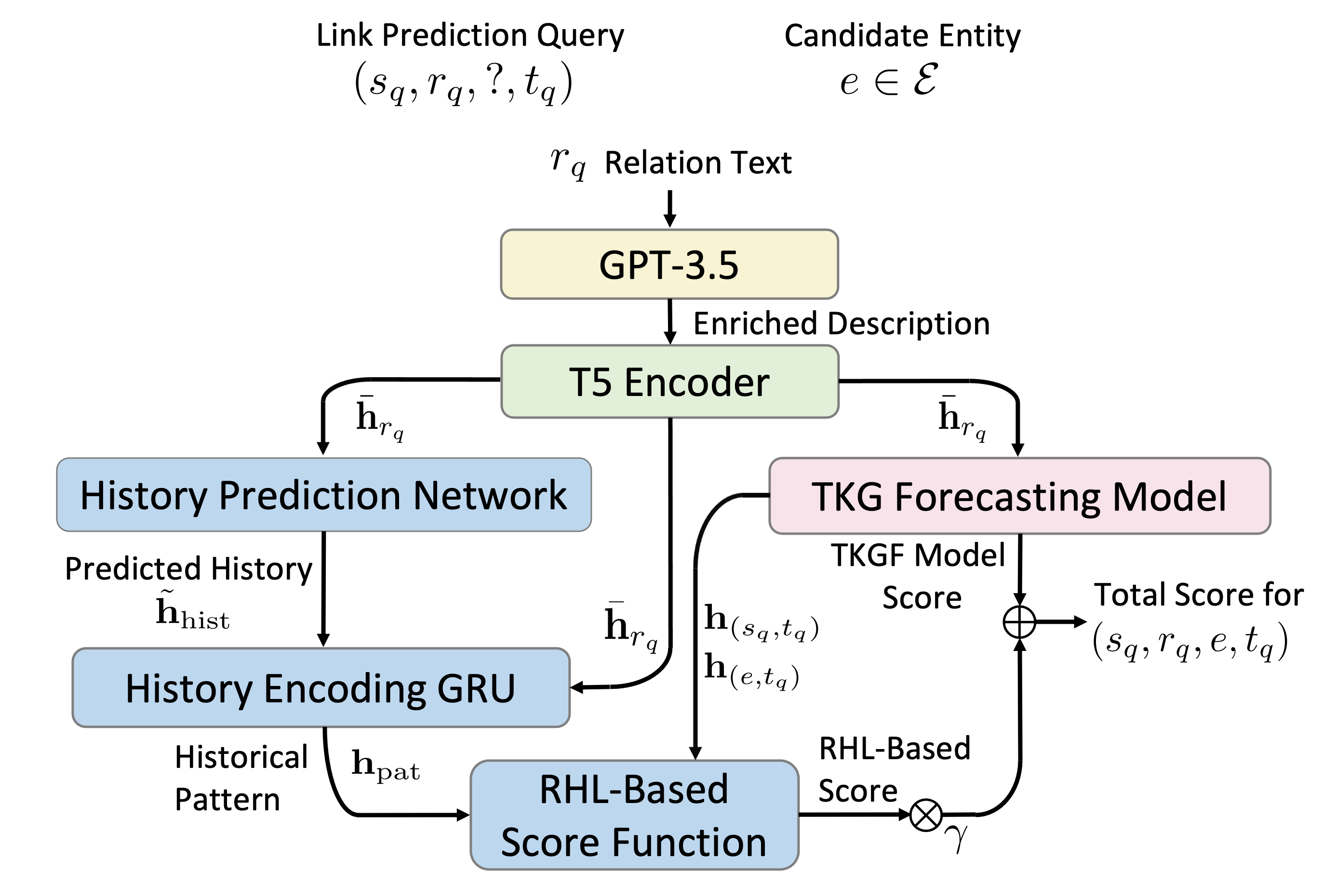}}
    \hfill
  \caption{Illustration of zrLLM-enhanced TKGF models. RHL-related components are marked in blue. RHL works differently in training and evaluation. During training, since we know both entities ($s,o$ in \ref{fig: model_train}) in the training fact, we can find the ground truth historical relations between them over time. We train a history prediction network (HPN) that aims to generate the relation history between two entities given their current relation ($r$). During evaluation, we directly use the trained HPN to infer the relation history. See Sec. \ref{sec: model} for details.
  }
  \label{fig: model structure} 
\end{figure*}
\section{zrLLM}
\label{sec: model}
zrLLM is coupled with TKGF models to enhance zero-shot ability. It uses GPT-3.5 to generate enriched relation descriptions (ERDs) based on the relation texts provided by TKG datasets. It further inputs the ERDs into the encoder of T5-11B and aligns its output to TKG embedding space. zrLLM also employs a relation history learner (RHL) to capture the temporal relation patterns based on the LLM-empowered relation representations. See Fig. \ref{fig: model structure} for illustration of zrLLM-enhanced TKGF models.
\subsection{Represent KG Relations with LLMs}
\paragraph{Generate Text Representations with ERDs.}
\label{sec: enrich description}
\begin{figure}
    \centering
    \includegraphics[width=0.75\columnwidth]{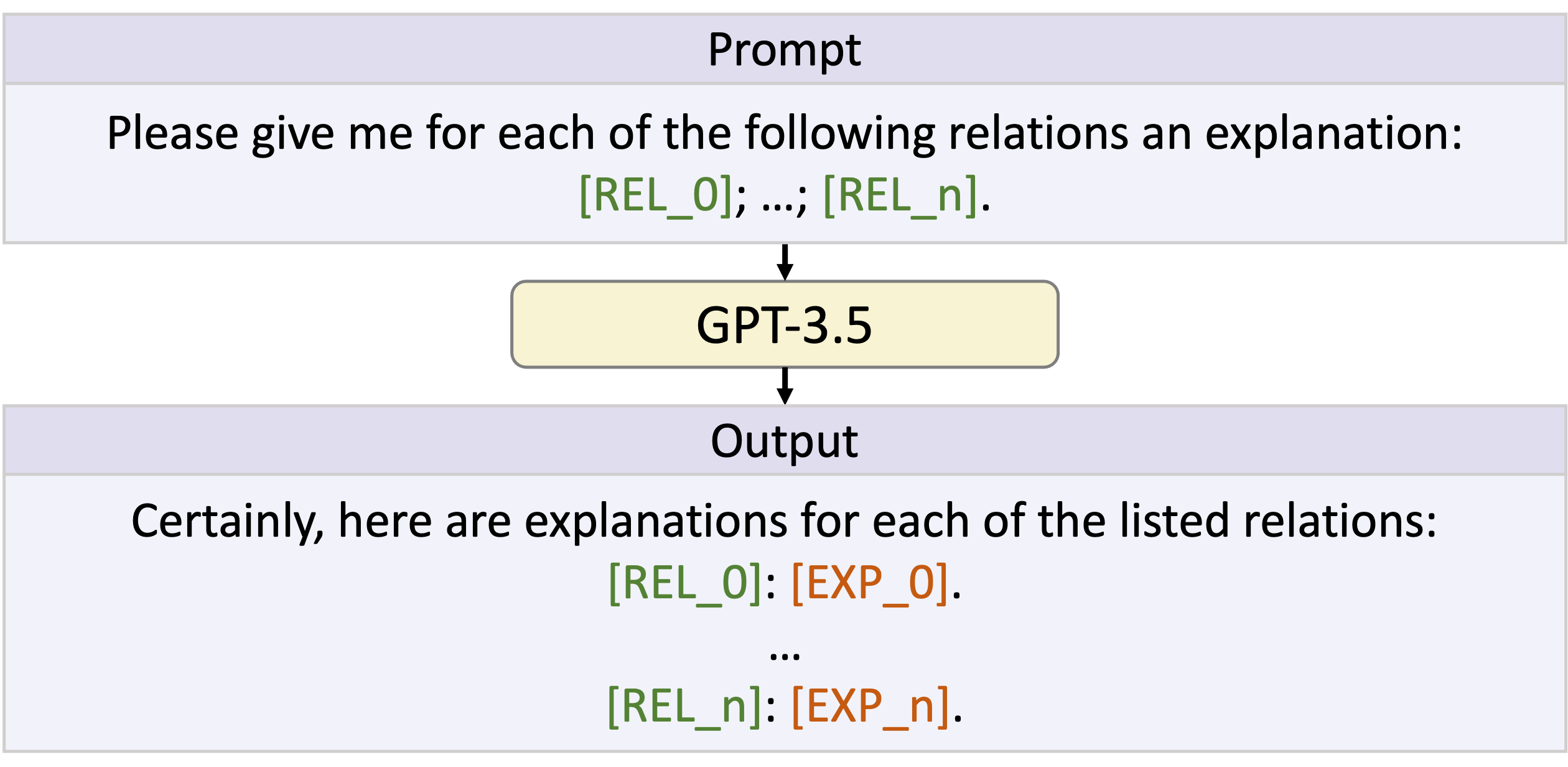}
    \caption{Prompting GPT-3.5 for ERDs. [REL\_$0$], ..., [REL\_$n$] are the dataset provided relation texts for a batch of $n$ KG relations. [EXP\_$0$], ..., [EXP\_$n$] are the LLM-generated explanations. [REL:\_$0$]: [EXP\_$0$], ..., [REL:\_$n$]: [EXP\_$n$] are taken as ERDs. See Appendix \ref{app: erd figure} for an expanded version of this figure.}
    \label{fig: prompt}
\end{figure}
We generate text representations with T5-11B based on the textual descriptions of KG relations. Since the relation texts provided by TKG datasets are short and concise, we use GPT-3.5\footnote{https://platform.openai.com/docs/model-index-for-researchers} to enrich them for more comprehensive semantics. 
Our prompt for description enrichment is depicted in Fig. \ref{fig: prompt}. 
For each relation, we treat the combination of its relation text and LLM-generated explanation as its ERD.
See Table \ref{tab: enrich example} for two enrichment examples.
\begin{table}[htbp]
\small
    \centering
    \resizebox{\columnwidth}{!}{
\begin{tabular}{c c} 
\toprule
\textbf{KG Relation Text}&\textbf{Enriched Relation Description}
\\ 
\midrule
\multirow{2}{*}{Engage in negotiation} & Engage in negotiation: This indicates a willingness to participate in discussions or
\\
& dialogues with the aim of reaching agreements or settlements on various issues.
\\
\midrule
\multirow{2}{*}{Praise or endorse} & Praise or endorse: This signifies a positive evaluation or approval of another entity's
\\
&  actions, policies, or behavior. It is a form of expressing support or admiration.
\\

\bottomrule
\end{tabular}
}
\caption{Relation description enrichment examples.}
\label{tab: enrich example}
\end{table}

We then input the ERDs into T5-11B. T5 is with an encoder-decoder architecture, where its encoder can be taken as a module that helps to understand the text input and the decoder is solely used for text generation. We take the output of T5-11B's encoder, i.e., the hidden representations, for our downstream task. Note that although ERDs are produced by GPT-3.5 who is trained with the corpus until the end of 2021, the representations used for TKGF are generated only with T5-11B, preventing information leak. Also, through our prompt, GPT-3.5 does not know our underlying task of TKGF. We manually check the ERDs generated by GPT-3.5 and make sure that GPT-3.5 generates relation explanations solely from the semantic perspective and no world knowledge is contained in its output.

\paragraph{Align Text Representations to TKG Embedding Space.}
\label{sec: alignment}
For each KG relation $r$, the T5-generated text representation is a parameter matrix $\Bar{\mathbf{H}}_{r} \in \mathbb{R}^{L\times d_w}$. $L$ is the length of the Transformers \cite{DBLP:conf/nips/VaswaniSPUJGKP17} in T5 and $d_w$ is the embedding size of each word output from T5 encoder. The $l^{\text{th}}$ row in $\Bar{\mathbf{H}}_{r}$ is the T5 encoded hidden representation $\mathbf{w}_l\in \mathbb{R}^{d_w}$ of the $l^{\text{th}}$ word in the enriched description. To align $\Bar{\mathbf{H}}_{r}$ to an embedding-based TKGF model, we first use a multi-layer perceptron (MLP) to map each $\mathbf{w}_l$ to the dimension of the TKGF model's relation representation.
\begin{equation}
\label{eq: align1}
\resizebox{0.6\columnwidth}{!}{%
$
    \mathbf{w}'_l = \text{MLP}(\mathbf{w}_l), \text{where}\  \mathbf{w}'_l\in \mathbb{R}^{d}.
$
}
\end{equation}
Then we learn a representation of $r$'s ERD $\Bar{\mathbf{h}}_{r}$ using a GRU.
\begin{equation}
\label{eq: align2}
\resizebox{0.7\columnwidth}{!}{%
$
\begin{aligned}
    &\Bar{\mathbf{h}}^{(l)}_{r} = \text{GRU}(\mathbf{w}'_l,\Bar{\mathbf{h}}^{(l-1)}_{r}); \ 
    \Bar{\mathbf{h}}^{(0)}_{r} = \mathbf{w}'_0,\\
    &\Bar{\mathbf{h}}_{r} = \Bar{\mathbf{h}}^{(L-1)}_{r}.
\end{aligned}
$
}
\end{equation}
$l \in [1, L-1]$. $\Bar{\mathbf{h}}_{r}$ contains semantic information from ERD, and therefore, we can view it as an LM-based relation representation. We substitute the relation representations of TKGF models with LM-based representations for semantics integration. 
Note that we fix the values of every $\Bar{\mathbf{H}}_{r}$ to keep the LLM-provided semantic information intact. This is because
we do not want the relation representations to lay excessive emphasis on the training data where zero-shot relations never appear. We want the models to maximally benefit from the semantic information for better generalization power. The textual descriptions of the relations with close meanings will show similar semantics. Since for each relation $r$, $\Bar{\mathbf{H}}_{r}$ is generated based on $r$'s ERD, the relations with close meanings will naturally lead to highly correlated text representations, building connections on top of the natural language space regardless of the observed TKG data. 
\subsection{Improving Text-to-Graph Alignment with Relation History Learner}
\label{sec: rhl}
As the relationship between two entities evolves through time, it follows certain temporal patterns. For example, the fact (\textit{China}, \textit{Sign} \textit{formal agreement}, \textit{Nicaragua},	2022-01-10) happens after (\textit{China}, \textit{Grant diplomatic recognition}, \textit{Nicaragua}, 2022-01-04), implying that an agreement will be signed after showing diplomatic recognition. These temporal patterns are entity-agnostic 
and can reflect the dynamic relationship between any two entities over time. To this end, we develop RHL, aiming to capture such patterns. RHL leverages the LLM-based relation representations for pattern modeling, which further promotes the alignment between the text and TKG embedding spaces.
Assume we have a training fact $(s, r, o, t)$, we search for the historical facts $\mathcal{G}_{s,o}^{<t}$ containing $s$ and $o$ before $t$, and group these facts according to their timestamps, i.e., $\mathcal{G}_{s,o}^{<t} = \{\mathcal{G}_{s,o}^{0}, ..., \mathcal{G}_{s,o}^{t-1}\}$. The searched facts with the same timestamp are put into the same group. For each group, we pick out the relations of all its facts and form a relation set, e.g., $\mathcal{R}_{s,o}^{0}$ is derived from $\mathcal{G}_{s,o}^{0}$. 
$s$ and $o$'s relationship at $t_i$ ($t_i \in [0, t-1]$) is computed with an aggregator
\begin{equation}
\resizebox{0.89\columnwidth}{!}{%
$
    \mathbf{h}^{t_i}_{s,o} = \sum_m a_m \Bar{\mathbf{h}}_{r_m}; \ 
    a_m = \text{softmax}(\Bar{\mathbf{h}}_{r_m}^\top \text{MLP}_{\text{agg}}(\Bar{\mathbf{h}}_{r})).
$
}
\end{equation}
$r_m \in \mathcal{R}_{s,o}^{t_i}$ denotes a relation bridging $s$ and $o$ at $t_i$. If $\mathcal{R}_{s,o}^{t_i} = \emptyset$, we set $\mathbf{h}^{t_i}_{s,o}$ to a dummy embedding $\mathbf{h}_\text{dum}$. 
To capture the historical relation dynamics, we use another GRU, i.e., $\text{GRU}_{\text{RHL}}$.
\begin{equation}
\label{eq: hist}
\resizebox{0.78\columnwidth}{!}{%
$
\begin{aligned}
    &\mathbf{h}_{\text{hist}}^{t_i} = \text{GRU}_{\text{RHL}}(\mathbf{h}^{t_i}_{s,o}, \mathbf{h}_{\text{hist}}^{t_i-1}); \ 
    \mathbf{h}_{\text{hist}}^{0} = \mathbf{h}^{0}_{s,o},\\
    &\mathbf{h}_{\text{hist}} = \mathbf{h}_{\text{hist}}^{t-1}.
\end{aligned}
$
}
\end{equation}
$\mathbf{h}_{\text{hist}}$ is taken as the encoded relation history until $t-1$. Note that during evaluation, TKGF asks models to predict the missing object of each LP query $(s_q, r_q, ?, t_q)$, which means we do not know which two entities should be used for historical fact searching\footnote{We can indeed couple $s_q$ with every candidate entity $e \in \mathcal{E}$ and search for their historical facts. But it requires huge computational resources and greatly harms model's scalability.}. To solve this problem, during training, we train another history prediction network (HPN) that aims to directly infer the relation history given the training fact relation $r$.
\begin{equation}
\label{eq: path pred}
\resizebox{0.55\columnwidth}{!}{%
$
    \Tilde{\mathbf{h}}_{\text{hist}} = \alpha \text{MLP}_{\text{hist}}(\Bar{\mathbf{h}}_{r}) + \Bar{\mathbf{h}}_{r}.
$
}
\end{equation}
Here, $\alpha$ is a hyperparameter scalar and $\text{MLP}_{\text{hist}}$ is an MLP. $\Tilde{\mathbf{h}}_{\text{hist}}$ is the predicted relation history given $r$. Since we want $\Tilde{\mathbf{h}}_{\text{hist}}$ to represent the ground truth relation history, we use a mean square error (MSE) loss to constrain it to be close to $\mathbf{h}_{\text{hist}}$.
\begin{equation}
\label{eq: mse}
\resizebox{0.5\columnwidth}{!}{%
$
    \mathcal{L}_{\text{hist}} = \text{MSE}(\Tilde{\mathbf{h}}_{\text{hist}}, \mathbf{h}_{\text{hist}}).
$
}
\end{equation}
In this way, during evaluation, we can directly use Eq. \ref{eq: path pred} to generate a meaningful $\Tilde{\mathbf{h}}_{\text{hist}}$ for further computation. Given $\Tilde{\mathbf{h}}_{\text{hist}}$, we do one more step in $\text{GRU}_{\text{RHL}}$ to capture the $r$-related relation pattern.
\begin{equation}
\label{eq: final path}
\resizebox{0.54\columnwidth}{!}{%
$
    \mathbf{h}_{\text{pat}} = \text{GRU}_{\text{RHL}}(\Bar{\mathbf{h}}_{r}, \Tilde{\mathbf{h}}_{\text{hist}}).
$
}
\end{equation}
$\mathbf{h}_{\text{pat}}$ can be viewed as a hidden representation containing comprehensive information of temporal relation patterns. Inspired by TuckER \cite{DBLP:conf/emnlp/BalazevicAH19}, we compute an RHL-based score for the training target $(s,r,o,t)$ as
\begin{equation}
\label{eq: rhl score}
\resizebox{0.85\columnwidth}{!}{%
$
    \phi((s,r,o,t)) = \mathcal{W} \times_1 \mathbf{h}_{(s, t)} \times_2 \mathbf{h}_{\text{pat}} \times_3 \mathbf{h}_{(o, t)},
$
}
\end{equation}
where $\mathcal{W} \in \mathbb{R}^{d \times d \times d}$ is a learnable core tensor and $\times_1, \times_2, \times_3$ are three operators indicating the tensor product in three different modes (details in \cite{DBLP:conf/emnlp/BalazevicAH19}). $\mathbf{h}_{(s, t)}$ and $\mathbf{h}_{(o, t)}$ are the time-aware entity representations of $s$ and $o$ computed by TKGF model, respectively. RHL-based score can be viewed as measuring how much two entities match the relation pattern generated by the relation history. We couple this score with the score computed by the original TKGF model $\phi'((s,r,o,t))$ and use the total score for LP.
\begin{equation}
\label{eq: total score}
\resizebox{0.89\columnwidth}{!}{%
$
    \phi_\text{total}((s,r,o,t)) = \phi'\left((s,r,o,t)) + \gamma \phi((s,r,o,t)\right).
$
}
\end{equation}
$\gamma$ is a hyperparameter. RHL enables models to make decisions by additionally considering the temporal relation patterns. Note that patterns are captured with LLM-empowered relation representations that contain rich semantic information. This guarantees RHL to generalize well to zero-shot relations. See App. \ref{app: rhl} for explanations.

\subsection{Parameter Learning and Evaluation}
We let zrLLM be co-trained with TKGF model. Assume $f$ is a TKGF model's loss function, e.g., cross-entropy, where $f$ takes a fact quadruple's score computed by model's score function $\phi'$ and returns a loss for this fact. We input the quadruple score computed with Eq. \ref{eq: total score} into $f$ to let TKGF models better learn the parameters in RHL.
\begin{equation}
\label{eq: tkgf loss}
\resizebox{0.72\columnwidth}{!}{%
$
    \mathcal{L}_\text{TKGF} = \frac{1}{|\mathcal{G}_\text{train}|}\sum_{\lambda \in \mathcal{G}_\text{train}}f(\phi_\text{total}(\lambda)),
$
}
\end{equation}
where $\lambda$ denotes a fact quadruple $(s, r, o, t) \in \mathcal{G}_\text{train}$ in the training set $\mathcal{G}_\text{train}$.
Besides, we also employ an additional binary cross-entropy loss $\mathcal{L}_\text{RHL}$ directly on the RHL-based score
\begin{equation}
\label{eq: RHL loss}
\resizebox{0.87\columnwidth}{!}{%
$
\begin{aligned}
        &\mathcal{L}_\text{RHL} = \frac{1}{N} \sum_{\lambda}\sum_{e \in \mathcal{E}} \mathcal{L}_\text{RHL}^{\lambda, e};\\
        &\mathcal{L}_\text{RHL}^{\lambda, e} = -y_{\lambda'} \log (\phi(\lambda')) - (1-y_{\lambda'})\log (1 - \phi(\lambda')).
\end{aligned}
$
}
\end{equation}
$N = |\mathcal{G}_\text{train}| \times |\mathcal{E}|$. $\lambda'$ is a perturbed fact by switching the object of $\lambda$ to any $e \in \mathcal{E}$ and $y_{\lambda'}$ is its label. If $\lambda' \in \mathcal{G}_\text{train}$, then $y_{\lambda'} = 1$, otherwise $y_{\lambda'} = 0$. Finally, we define the total loss $\mathcal{L}_\text{total}$ as
\begin{equation}
\label{eq: total loss}
   \mathcal{L}_\text{total} = \mathcal{L}_\text{TKGF} + \mathcal{L}_\text{hist} + \eta \mathcal{L}_\text{RHL}.
\end{equation}
$\eta$ is a hyperparameter deciding $\mathcal{L}_\text{RHL}$'s magnitude. Given our loss, we can also view RHL as a module that does a subtask during training. The subtask is to leverage the relation patterns encoded solely with LLM-based relation representations to perform TKG forecasting, which is parallel to the pipeline of the original TKGF model. This subtask training process helps to improve the embedding space alignment between text and graph representations. During evaluation, for each LP query $(s_q,r_q,?,t_q)$, we compute scores $\{\phi_\text{total}((s_q,r_q,e,t_q))\}|e\in \mathcal{E}\}$ and take the entity with maximum score as the predicted answer. We provide algorithms of training and evaluation in App. \ref{app: alg}.
\section{Experiments}
We give details of our new zero-shot TKGF datasets in Sec. \ref{sec: datasets}. In Sec. \ref{sec: main result}, we (1) do a comparative study to show how zrLLM improves TKGF models, (2) do ablation studies, (3) compare zrLLM with recent LM-enhanced TKGF models, and (4) do a case study to prove RHL's effectiveness. The implementation code and our proposed zero-shot datasets are in the following page: https://github.com/ZifengDing/zrLLM
\begin{table}[htbp]
    \centering
    \resizebox{\columnwidth}{!}{
\begin{tabular}{c c c c c c c c c c} \hline
\textbf{Dataset}&$|\mathcal{E}|$&$|\mathcal{R}|$&$|\mathcal{T}_{\text{train}}|$&$|\mathcal{T}_{\text{eval}}|$&$|\mathcal{R}_{se}|$&$|\mathcal{R}_{un}|$&$|\mathcal{G}_{\text{train}}|$&$|\mathcal{G}_{\text{valid}}|$&$|\mathcal{G}_{\text{test}}|$\\ 
\hline
ACLED-zero  & 621 & 23 & 20 & 11 & 9 & 14 & 2,118 & 931 & 146 \\ 
ICEWS21-zero  & 18,205 & 253 & 181 & 62 & 130 & 123 & 247,764 & 77,195 & 1,395  \\ 
ICEWS22-zero  & 999 & 248 & 181 & 62 & 93 & 155 & 171,013 & 47,784 & 1,956 \\ 
\hline
\end{tabular}
}
    \caption{Dataset statistics. Dataset timestamps consist of both training and evaluation timestamps, i.e., $\mathcal{T} = \mathcal{T}_{\text{train}} \cup \mathcal{T}_{\text{eval}}$, $\mathcal{T}_{\text{train}} \cap \mathcal{T}_{\text{eval}} = \emptyset$, $\text{max}(\mathcal{T}_{\text{train}}) < \text{min}(\mathcal{T}_{\text{eval}})$.
    }
\label{tab: data statistics}
\end{table}
\subsection{Datasets for Zero-Shot TKGF}
\label{sec: datasets}
As discussed in Sec. \ref{sec: TKG with LLM}, LM-enhanced TKGF models experience the risk of information leak. To exclude this concern, we construct new benchmark datasets on top of the facts happening after the publication date of T5-11B. We first construct two datasets ICEWS21-zero and ICEWS22-zero based on the Integrated Crisis Early Warning System (ICEWS) \cite{DVN/28075_2015} KB.
ICEWS21-zero contains the facts happening from 2021-01-01 to 2021-08-31, while all the facts in ICEWS22-zero happen from 2022-01-01 to 2022-08-31. Besides, we also construct another dataset ACLED-zero based on a newer KB: The Armed Conflict Location \& Event Data Project (ACLED) \cite{doi:10.1177/0022343310378914}.
Facts in ACLED-zero take place 
from 2023-08-01 to 2023-08-31. 
All the facts in all three datasets are based on social-political events described in English.
Inspired by \cite{DBLP:conf/akbc/MirtaheriR0MG21}, our dataset construction process consists of the following steps. (1) For each dataset, we first collect all the facts within the time period of interest from the associated KB and then sort them in the temporal order. (2) Then we split the collected facts into two splits, where the first split contains the facts for model training and the second one has all the facts for evaluation. Any fact from the evaluation split happens later than the maximum timestamp of all the facts from the training split. 
Since we are studying zero-shot relations, we exclude the facts in the evaluation split whose entities do not appear in the training split, to avoid the potential impact of unseen entities. (3) We compute the frequencies of all relations in the evaluation split, and set a frequency threshold (40 for ACLED-zero and ICEWS21-zero, 60 for ICEWS22-zero). (4) We take each relation whose frequency is lower than the threshold as a zero-shot relation, and treat every fact containing it in the evaluation split as zero-shot evaluation data $\mathcal{G}_{\text{test}}$. We exclude the facts associated with zero-shot relations from the training split to ensure that models cannot see these relations during training, and take the rest as the training set $\mathcal{G}_{\text{train}}$. The rest of facts 
in the evaluation split are taken as the regular evaluation data $\mathcal{G}_{\text{valid}}$. 
We do validation over $\mathcal{G}_{\text{valid}}$ and test over $\mathcal{G}_{\text{test}}$ because
we want to study how models perform over zero-shot relations when they reach the best performance over seen relations. 
See Table \ref{tab: data statistics} and App. \ref{app: dataset} for dataset statistics.
\begin{table*}[htbp]
    \centering
    \resizebox{\textwidth}{!}{
    \large\begin{tabular}{@{}lccc ccc c ccc ccc c ccc ccc c@{}}
\toprule
        \textbf{Datasets} & \multicolumn{7}{c}{\textbf{ACLED-zero}} & \multicolumn{7}{c}{\textbf{ICEWS21-zero}} & \multicolumn{7}{c}{\textbf{ICEWS22-zero}} \\
\cmidrule(lr){2-8} \cmidrule(lr){9-15} \cmidrule(lr){16-22}
        & \multicolumn{3}{c}{Zero-Shot Relations} & \multicolumn{3}{c}{Seen Relations} & Overall & \multicolumn{3}{c}{Zero-Shot Relations} & \multicolumn{3}{c}{Seen Relations} & Overall & \multicolumn{3}{c}{Zero-Shot Relations} & \multicolumn{3}{c}{Seen Relations} & Overall\\
\cmidrule(lr){2-4} \cmidrule(lr){5-7} \cmidrule(lr){8-8} \cmidrule(lr){9-11} \cmidrule(lr){12-14} \cmidrule(lr){15-15} \cmidrule(lr){16-18} \cmidrule(lr){19-21} \cmidrule(lr){22-22}
        \textbf{Model} & MRR & Hits@1 & Hits@10 & MRR & Hits@1 & Hits@10 & MRR  & MRR & Hits@1 & Hits@10 & MRR & Hits@1 & Hits@10 & MRR & MRR & Hits@1 & Hits@10 & MRR & Hits@1 & Hits@10 & MRR\\
\midrule 
        CyGNet 
        & 0.487 & 0.349  & \textbf{0.791}
        & \textbf{0.751}	& 0.663	& 0.903
        & 0.717
        & 0.120 & 0.046  & 0.270
        & 0.254 & \textbf{0.165}  & 0.432
        & 0.252
        & 0.211	& 0.098		& 0.459
        & \textbf{0.315}	& 0.198		& 0.540
        & 0.311
        
         \\
        CyGNet+
        & \textbf{0.533} & \textbf{0.418} & 0.753 
        & \textbf{0.751} & \textbf{0.664} & \textbf{0.906}
        & \textbf{0.723}
        & \textbf{0.201}	& \textbf{0.103}		& \textbf{0.415} 
        & \textbf{0.258}	& 0.162		& \textbf{0.447} & \textbf{0.257}
        & \textbf{0.286}	& \textbf{0.167}		& \textbf{0.542}
        & \textbf{0.315}	& \textbf{0.200}		& \textbf{0.545} & \textbf{0.314}
        
        \\
\midrule
        TANGO-T 
        & 0.052 & 0.021	& 0.101
        & 0.774	& 0.701	& 0.900
        & 0.681
        & 0.067 & 0.031	& 0.132
        & \textbf{0.283}	& \textbf{0.190}	& \textbf{0.470}
        & \textbf{0.279}
        & 0.092 & 0.042	& 0.187
        & \textbf{0.363}	& 0.250	& 0.579
        & 0.352
        
        \\
        TANGO-T+ 
        & \textbf{0.525}	& \textbf{0.393} & \textbf{0.764} 
        & \textbf{0.775}	& \textbf{0.702} & \textbf{0.901}
        & \textbf{0.743}
        & \textbf{0.216}	& \textbf{0.125}  & \textbf{0.395} 
        & 0.280	& 0.186 & 0.466 & \textbf{0.279}
        & \textbf{0.326}	& \textbf{0.198} & \textbf{0.578} 
        & \textbf{0.363}	& \textbf{0.251} & \textbf{0.585} & \textbf{0.362}
        
        \\
\midrule       
       TANGO-D
       & 0.021		& 0.003	& 0.049
       & \textbf{0.777}	& \textbf{0.701} & \textbf{0.907} 
       & 0.679
       & 0.012		& 0.005		& 0.023
       & 0.266	& \textbf{0.178}  & 0.439
       & 0.261
       & 0.011		& 0.002		& 0.018
       & \textbf{0.350}	& 0.227 & 0.569
       & 0.337
        
        \\
       TANGO-D+
       & \textbf{0.491}	& \textbf{0.348}		& \textbf{0.791} 
       & 0.760	& 0.678 & 0.901 
       & \textbf{0.725}
       & \textbf{0.212}	& \textbf{0.122}		& \textbf{0.400} 
       & \textbf{0.268}	& 0.175  & \textbf{0.453}   & \textbf{0.267}
       & \textbf{0.311}	& \textbf{0.186}	& \textbf{0.574}	
        & \textbf{0.350}	& \textbf{0.239}	& \textbf{0.570} & \textbf{0.348}
        
        \\
\midrule
        RE-GCN
        & 0.441	& 0.332	& 0.718
        & 0.730	& \textbf{0.653} & 0.865 
        & 0.693
        & 0.200	& 0.104	& 0.379
        & 0.277	& 0.185 & \textbf{0.456} 
        & 0.276
        & 0.280	& 0.162	& \textbf{0.616}
        & 0.354	& 0.243 & 0.567
        & 0.351
        
        \\
        RE-GCN+ 
        & \textbf{0.529}	& \textbf{0.393} & \textbf{0.784} 
        & \textbf{0.731}	& 0.650 & \textbf{0.876}
        & \textbf{0.705}
        & \textbf{0.214}	& \textbf{0.117} & \textbf{0.406} 
        & \textbf{0.280}	& \textbf{0.188} & \textbf{0.456} & \textbf{0.279}
        & \textbf{0.324}	& \textbf{0.194} & 0.595
        & \textbf{0.357}	& \textbf{0.244} & \textbf{0.573} & \textbf{0.356}
        
        \\
\midrule
        TiRGN
        & 0.478	& 0.330	& 0.745
        & \textbf{0.754}	& 0.678	& \textbf{0.886}
        & 0.718
        & 0.189	& 0.101	& 0.368
        & 0.275	& 0.182	& 0.457
        & 0.273
        & 0.299	& 0.169	& 0.570
        & 0.352	& 0.239	& 0.575
        & 0.350
        \\
        TiRGN+ 
        & \textbf{0.548}	& \textbf{0.436}	& \textbf{0.750}	
        & \textbf{0.754}	& 0.679	& 0.885
        & \textbf{0.727}
        & \textbf{0.221}	& \textbf{0.130}	& \textbf{0.410}	
        & \textbf{0.279}	& \textbf{0.185}	& \textbf{0.464} & \textbf{0.278}
        & \textbf{0.333}	& \textbf{0.203}	& \textbf{0.602}	
        & \textbf{0.353}	& \textbf{0.240}	& \textbf{0.577} & \textbf{0.352}
        \\
\midrule
RETIA 
        & 0.499	& 0.360	& 0.795	
        & 0.782	& 0.701	& 0.924
        & 0.745
        &&&
        \multicolumn{3}{c}{\multirow{2}{*}{\text{>> 120 Hours Timeout}}}
        &&
        & 0.302	& 0.166	& 0.566	
        & 0.356	& 0.245	& 0.577
        & 0.354
        \\
        RETIA+ 
        & \textbf{0.557}	& \textbf{0.408}	& \textbf{0.814}	
        & \textbf{0.783}	& \textbf{0.703}	& \textbf{0.925}
        & \textbf{0.754}
        & &&&&&&
        & \textbf{0.331}	& \textbf{0.201}	& \textbf{0.597}	
        & \textbf{0.358}	& \textbf{0.247}	& \textbf{0.578} & \textbf{0.357}
        \\
\midrule
        CENET
        & 0.419	& 0.297	& 0.593	
        & 0.753	& 0.682	& 0.869
        & 0.710
        & 0.205	& 0.101	& 0.411	
        & 0.288	& 0.196	& 0.468
        & 0.287
        & 0.270	& 0.134	& 0.544	
        & 0.379	& 0.268	& 0.599
        & 0.375
        \\
        CENET+ 
        & \textbf{0.591}	& \textbf{0.451}	& \textbf{0.844}	
        & \textbf{0.779}	& \textbf{0.692}	& \textbf{0.912}
        & \textbf{0.755}
        & \textbf{0.335}	& \textbf{0.162}	& \textbf{0.659}		
        & \textbf{0.396}	& \textbf{0.239}	& \textbf{0.688} & \textbf{0.395}
        & \textbf{0.564}	& \textbf{0.432}	& \textbf{0.801}	
        & \textbf{0.571}	& \textbf{0.451}	& \textbf{0.773} & \textbf{0.570}
        \\
\bottomrule
    \end{tabular}}
\caption{LP results.
    The best results between each baseline and its zrLLM-enhanced version (model name with "+") are marked in bold. TANGO-T and TANGO-D denote TANGO with TuckER \cite{DBLP:conf/emnlp/BalazevicAH19} and Distmult \cite{DBLP:journals/corr/YangYHGD14a}, respectively. RETIA cannot be trained before 120 hours timeout on ICEWS21-zero. Complete results with Hits@3 are presented in App. \ref{app: complete main results}.}
\label{tab: tid LP results all datasets}
\end{table*}
\subsection{Experimental Setup}
\label{sec: exp setup}
\paragraph{Training and Evaluation for Zero-Shot TKGF.}
All TKGF models are trained on $\mathcal{G}_{\text{train}}$. We take the model checkpoint achieving the best validation result on $\mathcal{G}_{\text{valid}}$ as the best model checkpoint, and report their test result on $\mathcal{G}_{\text{test}}$ to study the zero-shot inference ability. To keep zero-shot relations "always unseen" during the whole test process, we constrain all models to do LP only based on the training set as several popular TKGF methods, e.g., RE-GCN \cite{DBLP:conf/aaai/ZhuCFCZ21}. Some TKGF models, e.g., TiRGN \cite{DBLP:conf/ijcai/LiS022}, allow using the ground truth TKG data until the LP query timestamp, including the facts in evaluation sets. This will violate the zero-shot setting because every unseen relation will occur multiple times in the evaluation data and is no longer zero-shot after models observe any fact of it. We prevent them from observing evaluation data to maintain the zero-shot setting. See App. \ref{app: zero setting} for explanation.
Note that $\mathcal{G}_{\text{valid}}$ and $\mathcal{G}_{\text{test}}$ share the same time period. This is because \textbf{we want to make sure that zrLLM can enhance zero-shot reasoning and simultaneously maintain TKGF models' performance on the facts with seen relations. Improving zero-shot inference ability at the cost of sacrificing too much performance over seen relations is undesired.}
\paragraph{Baselines and Evaluation Metrics.}
We consider seven recent embedding-based TKGF methods as baselines, i.e., CyGNet \cite{DBLP:conf/aaai/ZhuCFCZ21}, TANGO-TuckER/Distmult \cite{DBLP:conf/emnlp/HanDMGT21}, RE-GCN \cite{DBLP:conf/sigir/LiJLGGSWC21}, TiRGN \cite{DBLP:conf/ijcai/LiS022}, CENET \cite{DBLP:conf/aaai/XuO0F23} and RETIA \cite{DBLP:conf/icde/Liu0X0023}. We couple them with zrLLM and show their improvement in zero-shot relational learning on TKGs (implementation details in App. \ref{app: implementation details}).
We employ two evaluation metrics, i.e., mean reciprocal
rank (MRR) and Hits@1/3/10. See App. \ref{app: ablation} for detailed definitions. As suggested in \cite{DBLP:conf/pkdd/GastingerSSSS23}, we use the time-aware filtering setting \cite{DBLP:conf/iclr/HanCMT21} for fairer evaluation.
\subsection{Comparative Study and Further Analysis}
\label{sec: main result}
\paragraph{Comparative Study.}
We report the LP results of all baselines and their zrLLM-enhanced versions in Table \ref{tab: tid LP results all datasets}. We have two findings: (1) zrLLM greatly helps TKGF models in forecasting the facts with unseen zero-shot relations. (2) In most cases, zrLLM even improves models in predicting the facts with seen relations. 
The zrLLM-enhanced models whose performance drops over seen relations still achieve better overall performance.
These findings prove that embedding-based TKGF models benefit from the semantic information extracted from LLMs, especially when they are dealing with zero-shot relations. 
\paragraph{Ablation Study.}
\label{sec: further analysis}
We conduct ablation studies from three aspects. 
(1) First, we directly input the dataset provided relation texts into T5-11B encoder, ignoring the relation explanations generated by GPT-3.5. From Table \ref{tab: ablation} (-ERD), we observe that in most cases, models' performance drops on the facts with both seen and zero-shot relations,
which proves the usefulness of ERDs.
(2) Next, we remove the RHL from all zrLLM-enhanced models. From Table \ref{tab: ablation} (-RHL), we find that all the considered TKGF models can benefit from RHL, especially CENET. 
(3) We switch T5-11B to T5-3B to see the impact of LM size on zrLLM. 
We observe from Table \ref{tab: ablation} that decreasing the size of T5 harms model performance. 
This proves that using larger scale LMs can provide semantic information of higher quality, and can be more beneficial to downstream TKGF (whether zero-shot or not). 
\begin{table}[htbp]
    \centering
    \resizebox{0.95\columnwidth}{!}{
    \large\begin{tabular}{@{}lccc ccc ccc@{}}
\toprule
        \textbf{Datasets} & \multicolumn{3}{c}{\textbf{ACLED-zero}} & \multicolumn{3}{c}{\textbf{ICEWS21-zero}} & \multicolumn{3}{c}{\textbf{ICEWS22-zero}} \\
        & \multicolumn{3}{c}{MRR} & \multicolumn{3}{c}{MRR} & \multicolumn{3}{c}{MRR} \\
\cmidrule(lr){2-4} \cmidrule(lr){5-7} \cmidrule(lr){8-10}
        \textbf{Model} & Zero & Seen & Overall & Zero & Seen & Overall & Zero & Seen & Overall\\
\midrule 
        CyGNet+ 
        & \textbf{0.533} 
        & 0.751  
        & \textbf{0.723}
        & \textbf{0.201} 	
        & \textbf{0.258} 	
        & \textbf{0.257}
        & \textbf{0.286} 
        & \textbf{0.315} 
        & \textbf{0.314}
         \\
        \ - ERD
        & 0.502	
        & 0.748
        & 0.716
        & 0.198	
        & 0.252	
        & 0.251
        & 0.250		
        & 0.314	
        & 0.311
        \\
        \ - RHL
        & 0.503		
        & \textbf{0.752}
        & 0.720
        & 0.199	
        & 0.256	
        & 0.255
        & 0.268	
        & 0.297		
        & 0.296
        \\
        \ T5-3B
        & 0.511	
        & \textbf{0.752}	
        & 0.721
        & 0.117	
        & 0.204	
        & 0.202
        & 0.257		
        & \textbf{0.315}	
        & 0.313
        \\
\midrule
        TANGO-T+
        & 0.525	
        & \textbf{0.775}	
        & \textbf{0.743}
        & \textbf{0.216}	
        & \textbf{0.280}	
        & \textbf{0.279}
        & \textbf{0.326}		
        & \textbf{0.363}		
        & \textbf{0.362}
        \\
        \ - ERD
        & 0.533		
        & 0.772
        & 0.741
        & 0.214	
        & \textbf{0.280}		
        & \textbf{0.279}
        & 0.320	
        & 0.362		
        & 0.360
        \\
        \ - RHL
        & 0.506	
        & 0.755	
        & 0.740
        & 0.213		
        & 0.277		
        & 0.276
        & 0.309	
        & \textbf{0.363}	
        & 0.361
        \\
        \ T5-3B
        & \textbf{0.544}		
        & 0.771
        & 0.742
        & 0.206
        & 0.274	
        & 0.273
        & 0.323
        & 0.359
        & 0.358
        \\
\midrule       
       TANGO-D+
        & \textbf{0.491}	
        & \textbf{0.760}	
        & \textbf{0.725}
        & \textbf{0.212}		
        & \textbf{0.268}	
        & \textbf{0.267}
        & \textbf{0.311}		
        & \textbf{0.350}		
        & \textbf{0.348}
        \\
       \ - ERD
        & \textbf{0.491}
        & 0.702	
        & 0.675
        & 0.205	
        & 0.267		
        & 0.266
        & 0.285	
        & 0.328	
        & 0.326
        \\
        \ - RHL
        & 0.490
        & 0.725	
        & 0.695
        & 0.197	
        & 0.224		
        & 0.224
        & 0.296		
        & 0.324	
        & 0.323
        \\
        \ T5-3B
        & 0.490		
        & 0.701	
        & 0.674
        & 0.204	
        & 0.223		
        & 0.222
        & 0.308	
        & 0.284
        & 0.285
        \\
\midrule
        RE-GCN+
        & \textbf{0.529}		
        & \textbf{0.731}	
        & \textbf{0.705}
        & \textbf{0.214}		
        & \textbf{0.280}		
        & \textbf{0.279}
        & \textbf{0.324}	
        & \textbf{0.357}	
        & \textbf{0.356}
        \\
        \ - ERD
        & 0.489	
        & 0.730	
        & 0.699
        & 0.211		
        & 0.277	
        & 0.276
        & 0.294	
        & 0.354	
        & 0.352
        \\
        \ - RHL
        & 0.519	
        & 0.726
        & 0.699
        & 0.213	
        & 0.277	
        & 0.276
        & 0.317		
        & 0.350	
        & 0.349
        \\
        \ T5-3B
        & 0.504
        & 0.721
        & 0.693
        & 0.211		
        & 0.259	
        & 0.258
        & 0.301
        & 0.354
        & 0.352
        \\
\midrule
        TiRGN+
        & \textbf{0.548}
        & \textbf{0.754}	
        & \textbf{0.727}
        & \textbf{0.221}		
        & \textbf{0.279}	
        & \textbf{0.278}
        & \textbf{0.333}	
        & \textbf{0.353}		
        & \textbf{0.352}
        \\
        \ - ERD
        & 0.480		
        & 0.747
        & 0.713
        & 0.211		
        & 0.275		
        & 0.274
        & 0.282	
        & \textbf{0.353}	
        & 0.350
        \\
        \ - RHL
        & 0.515		
        & 0.752	
        & 0.721
        & 0.215		
        & 0.277		
        & 0.276
        & 0.320		
        & 0.350	
        & 0.349
        \\
        \ T5-3B
        & 0.498
        & 0.749
        & 0.717
        & 0.208	
        & 0.271
        & 0.270
        & 0.325
        & 0.345
        & 0.344
        \\
\midrule
        RETIA+
        & \textbf{0.557}		
        & \textbf{0.783}
        & \textbf{0.754}
        &
                \multicolumn{3}{c}{\multirow{4}{*}{\text{>> 120 Hours Timeout}}}
        
        & \textbf{0.331}	
        & \textbf{0.358}	
        & \textbf{0.357}
        \\
        \ - ERD
        & 0.519	
        & 0.777	
        & 0.744
        &&
        &
        & 0.292
        & 0.354	
        & 0.352
        \\
        \ - RHL
        & 0.529	
        & 0.782	
        & 0.749
        &&
        &
        & 0.318		
        & 0.357	
        & 0.355
        \\
        \ T5-3B
        & 0.512	
        & 0.776
        & 0.742
        &&
        &
        & 0.330
        & 0.353	
        & 0.352
        \\
\midrule
        CENET+
        & \textbf{0.591}		
        & \textbf{0.779}	
        & \textbf{0.755}
        & \textbf{0.335}	
        & \textbf{0.396}	
        & \textbf{0.395}
        & \textbf{0.564}		
        & \textbf{0.571}		
        & \textbf{0.570}
        \\
        \ - ERD
        & 0.526		
        & 0.737	
        & 0.710
        & 0.321	
        & 0.374	
        & 0.373
        & 0.542	
        & 0.570	
        & 0.568
        \\
        \ - RHL
        & 0.445	
        & 0.754	
        & 0.714
        & 0.232	
        & 0.290	
        & 0.289
        & 0.295	
        & 0.370	
        & 0.367
        \\
        \ T5-3B
        & 0.568	
        & 0.736
        & 0.714
        & 0.303	
        & 0.330	
        & 0.329
        & 0.550	
        & 0.555
        & 0.554
        \\
\bottomrule
    \end{tabular}}
\caption{Ablation study (complete results in App. \ref{app: ablation}).}
\label{tab: ablation}
\end{table}
\begin{figure*}[htbp]
\centering

\subfloat[]{%
  \includegraphics[width=0.45\textwidth]{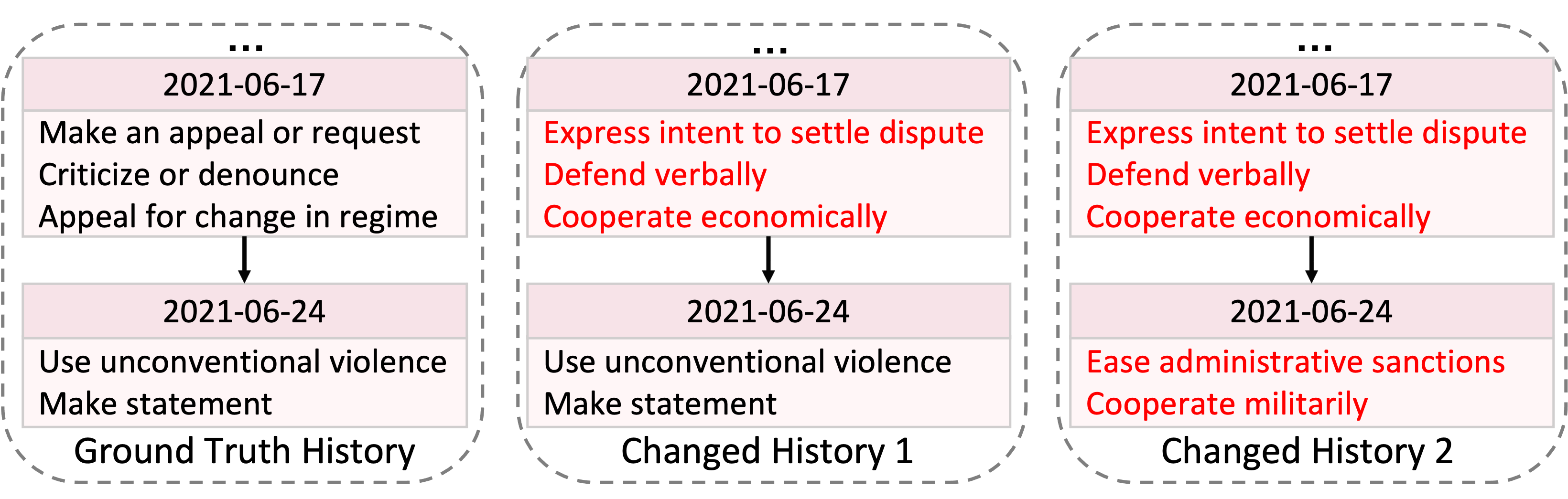}%
  \label{fig: case}%
}\hfill
\subfloat[]{%
  \includegraphics[width=0.38\textwidth]{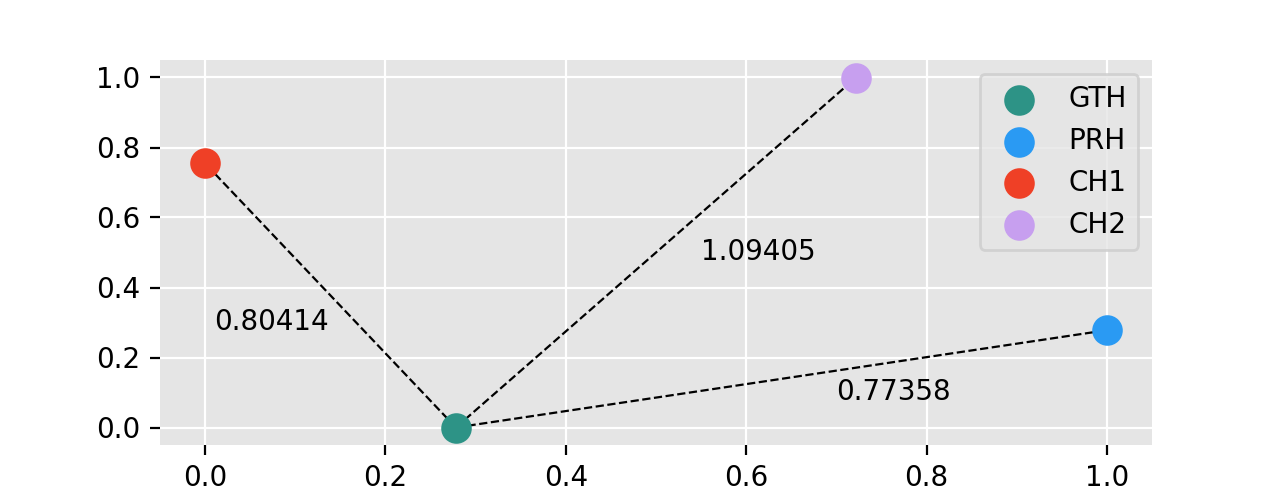}%
  \label{fig: case_tsne}%
}\hfill
\subfloat[]{%
  \includegraphics[width=0.15\textwidth]{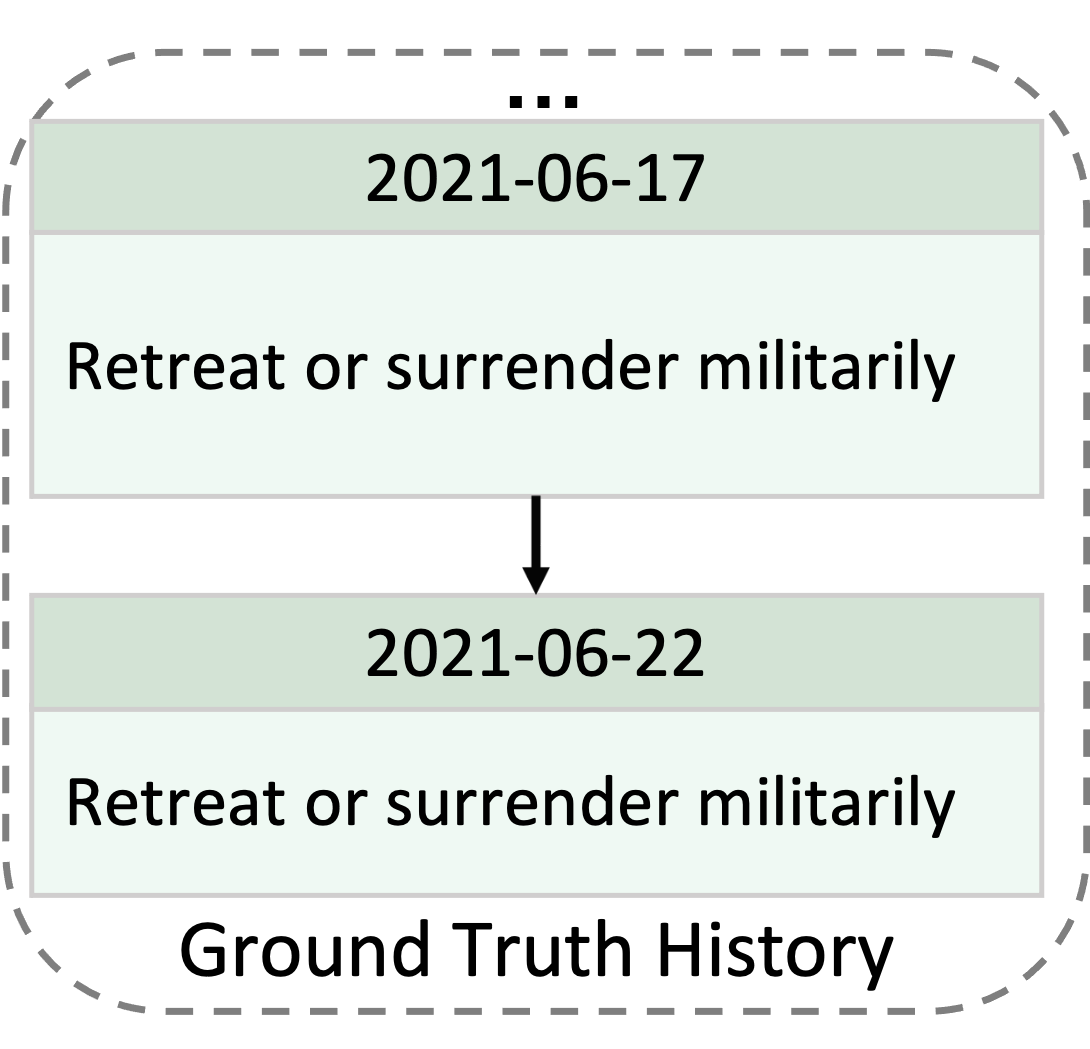}%
  \label{fig: case2}%
}
\caption{(a) Ground truth and changed relation histories between \textit{United States} and \textit{African Union}. Changed relations are marked in red. Only the histories nearest to 2021-07-03 are shown. (b) t-SNE of encoded GTH, CH1, CH2 (computed with Eq. \ref{eq: hist}), and predicted history PRH. Numbers beside dashed lines denote point distances (L2 norm). (c) Ground truth relation histories between \textit{United States} and \textit{Afghanistan}.}
\end{figure*}
\paragraph{Compare with Previous LM-Enhanced Model.}
We benchmark two recent LM-enhanced TKGF models PPT \cite{DBLP:conf/acl/XuLPJP23} and ICL + GPT-NeoX-20B \cite{DBLP:journals/corr/abs-2305-10613,DBLP:journals/corr/abs-2204-06745} (Table \ref{tab: previous lm}). PPT finetunes BERT for TKGF. 
We find that although PPT achieves strong zero-shot results, 
it is beaten by several zrLLM-enhanced models. 
This proves that aligning language space to TKGF is helpful for zero-shot relational learning and LMs with larger size can be more contributive. ICL shows inferior results. This proves that without finetuning or alignment, LLMs are unable to optimally solve TKGF. zrLLM not only benefits from a large LM but also enables efficient alignment from language to TKG embedding space, which leads to superior performance. See App. \ref{app: previous LM} for further discussion.
\begin{table}[htbp]
    \centering
    \resizebox{0.95\columnwidth}{!}{
    \large\begin{tabular}{@{}lccc ccc ccc@{}}
\toprule
        \textbf{Datasets} & \multicolumn{3}{c}{\textbf{ACLED-zero}} & \multicolumn{3}{c}{\textbf{ICEWS21-zero}} & \multicolumn{3}{c}{\textbf{ICEWS22-zero}} \\
        & \multicolumn{3}{c}{MRR} & \multicolumn{3}{c}{MRR} & \multicolumn{3}{c}{MRR} \\
\cmidrule(lr){2-4} \cmidrule(lr){5-7} \cmidrule(lr){8-10}
        \textbf{Model} & Zero & Seen & Overall & Zero & Seen & Overall & Zero & Seen & Overall\\
\midrule 
        PPT
        & 0.532	
        & 0.782	
        & 0.748
        & 0.212	
        & 0.269
        & 0.268
        & 0.323		
        & 0.332
        & 0.331
        \\
        ICL
        & 0.537
        & 0.736
        & 0.709
        & 0.156	
        & 0.178
        & 0.177
        & 0.255
        & 0.229		
        & 0.230
        \\
\bottomrule
    \end{tabular}}
\caption{PPT and ICL performance. Implementation details and complete results in App. \ref{app: implementation details PPT icl} and \ref{app: previous LM}.}
\label{tab: previous lm}
\end{table}
\paragraph{Case Study of RHL}
We do a case study to show: (1) RHL's HPN is able to capture ground truth relation history (GTH). (2) By capturing temporal relation patterns, RHL helps for better zero-shot TKGF.
We ask zrLLM-enhanced CENET to predict the missing object of the test query $q = (s_q, r_q, ?, t_q) = $ (\textit{United States}, \textit{Reduce or stop military assistance}, $?$, 2021-07-03) (answer is $o_q =$ \textit{African Union}) taken from ICEWS21-zero. The GTH of $s_q$ and $o_q$ (Fig. \ref{fig: case}, left) shows a pattern indicating their recent worsening relationship. It can serve as a clue in LP over $q$ because it can be viewed as a "cause" to the query relation $r_q$ which also implies a negative relationship. In other words, the entities with a worsening historical relationship are more likely to be connected with a relation showing their bad relationship currently. Since RHL uses HPN to infer GTH during test, we wish to study whether HPN can achieve reasonable inference to support LP. Based on GTH, we first change all three relations on 2021-06-17 to randomly sampled positive relations seen in the training data and form a changed history 1 (CH1, Fig. \ref{fig: case}, middle). Then we further modify the relations on 2021-06-24 in the same way and form a changed history 2 (CH2, Fig. \ref{fig: case}, right). We use Eq. \ref{eq: hist} to encode GTH, CH1, CH2, and visualize them together with the predicted history (PRH) computed with HPN by using t-SNE \cite{JMLR:v9:vandermaaten08a} in Fig. \ref{fig: case_tsne}. We find that PRH is the closest to GTH and CH1 is closer than CH2 to GTH. The reason why CH2 is much farther from GTH is that CH2 changes more negative relations to positive, greatly changing the semantic meaning stored in GTH. CH1 only introduces changes on 2021-06-17, making it less deviated from GTH. HPN takes the $r_q$ and can keep PRH close to GTH, making zrLLM able to maximally capture the temporal patterns indicated by GTH, while preventing the scalability problem incurred by searching relation histories of all candidate entities. 
By using RHL, the zrLLM-enhanced CENET can correctly predict $o_q$, while the model without RHL takes $o' = $ \textit{Afghanistan} as the predicted answer. We present the nearest GTH between $s_q$ and $o'$ in Fig. \ref{fig: case2} and find that it indicates a positive relationship which is unlikely to cause $r_q$ right after.
During training, RHL learns patterns and matches entity pairs with them (Eq. \ref{eq: rhl score}). This enables RHL to exclude the entities that do not fit into the learned patterns from the answer set and make more accurate predictions.
\section{Conclusion}
We study zero-shot relational learning in TKGF and design an LLM-empowered approach, i.e., zrLLM. zrLLM extracts the semantic information of KG relations from LLMs and introduces it into TKG representation learning. It also uses an RHL module to capture the temporal relation patterns for better reasoning. We couple zrLLM with several embedding-based TKGF models and find that zrLLM provides huge help in forecasting the facts with zero-shot relations, and moreover, it maintains models' performance over seen relations.
\section{Limitations}
Our limitations can be summarized as follows.
First, zrLLM is developed only for enhancing embedding-based TKG forecasting methods. It is not directly applicable to the rule-based methods, e.g., TLogic. Besides, relation history learner inevitably increases model's training and evaluation time since relation patterns are learned with GRUs where recurrent computations are performed along the time axis. More GPU memory is also required for storing relation histories. This hinders the efficiency of zrLLM-enhanced models compared with the original baselines. In the future, we will explore how to generalize our proposed method to rule-based models and try to improve model efficiency. We will also try to experiment zrLLM on more TKG forecasting methods and study whether we can benefit more of them.

\section*{Acknowledgments}
This research has been partially funded by the Munich Center for
Machine Learning and supported by the Federal
Ministry of Education and Research and the State
of Bavaria.
The authors thank the International Max Planck Research School for Intelligent Systems (IMPRS-IS) for supporting Bo Xiong. 
The research has also been partially funded by Deutsche Forschungsgemeinschaft (DFG, German Research Foundation) under Germany’s Excellence Strategy - EXC 2075 - 390740016, the Stuttgart Center for Simulation Science (SimTech), and the Bundesministerium für Wirtschaft und Energie (BMWi), grant aggrement No. 01MK20008F. 
\bibliography{custom}

\appendix
\section{Detailed Illustration of Prompt for GPT-3.5}
\label{app: erd figure}
\begin{figure*}
    \centering
    \includegraphics[width=0.7\textwidth]{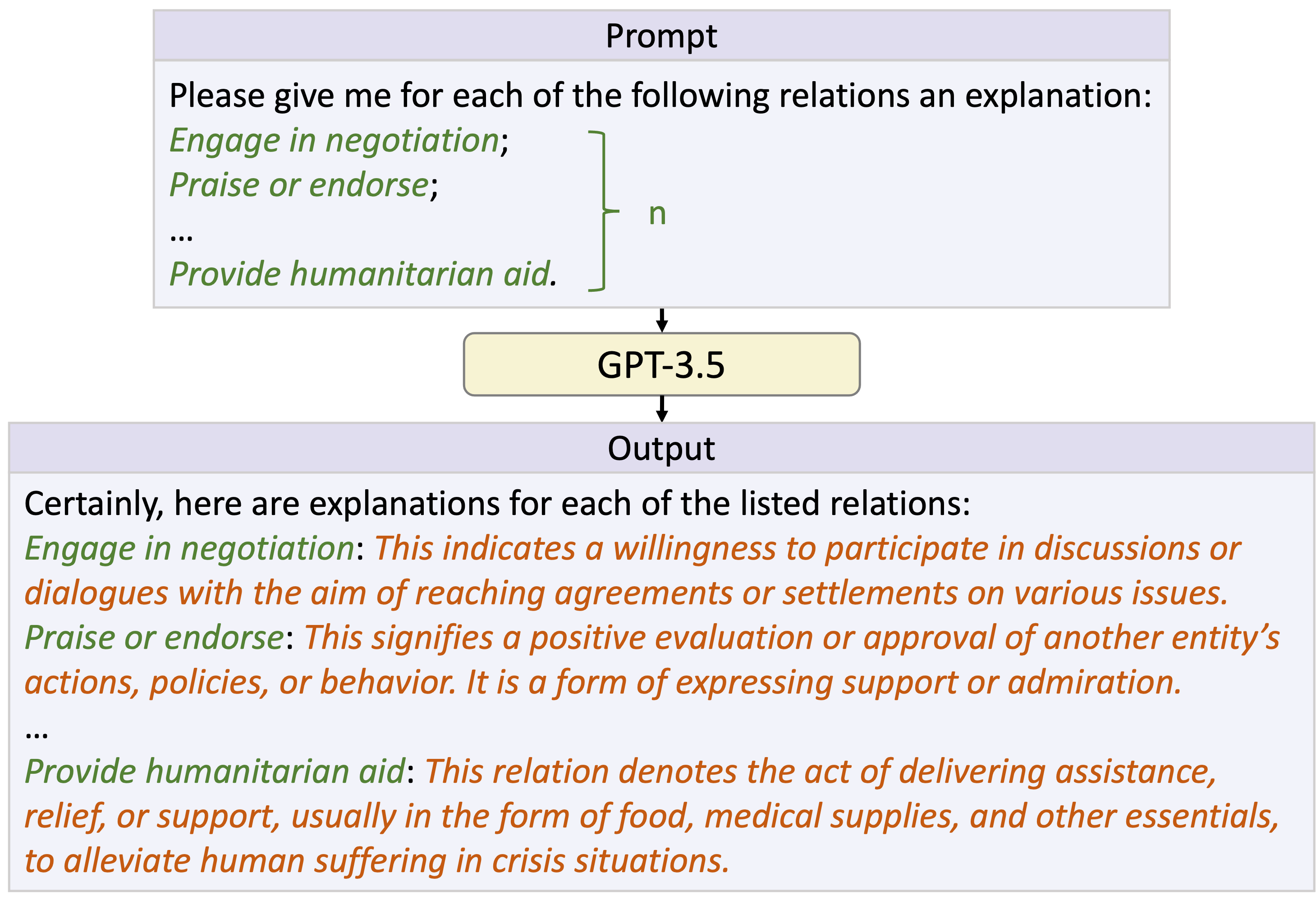}
    \caption{Prompting GPT-3.5 for ERDs. The green texts are the short relation texts provided in the original datasets. The orange texts are the generated relation explanations from GPT-3.5.}
    \label{fig: prompt detail}
\end{figure*}

We give a detailed illustration of our prompt for producing ERDs with GPT-3.5 in Fig. \ref{fig: prompt detail}. For every batch of $n$ relations, we incorporate their dataset-provided texts into our prompt to generate their enriched descriptions. 

\section{Further Details of Zero-Shot Datasets}
\label{app: dataset}
\begin{figure*}[t]
    \centering
  \subfloat[ACLED-zero.\label{fig: acled_freq}]{%
       \includegraphics[width=0.33\textwidth]{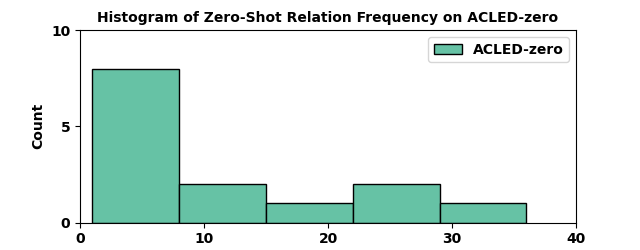}}
    \hfill
  \subfloat[ICEWS21-zero.\label{fig: 21_freq}]{%
        \includegraphics[width=0.33\textwidth]{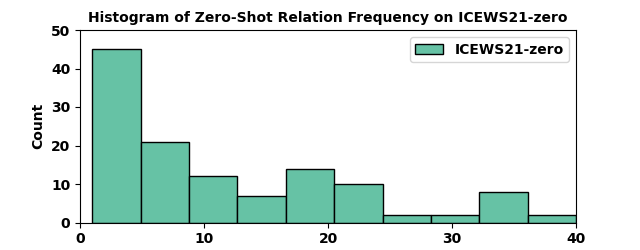}}
    \hfill
  \subfloat[ICEWS22-zero.\label{fig: 22_freq}]{%
        \includegraphics[width=0.33\textwidth]{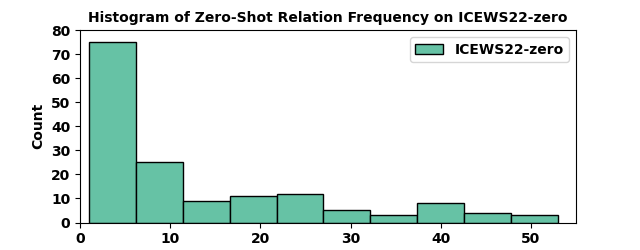}}
    \hfill
  \caption{Zero-shot Relation frequency on all zero-shot TKGF datasets. Horizontal axis denotes the appearance times, i.e., frequency. Vertical axis denotes the number of relations.}
  \label{fig: data comparison} 
\end{figure*}

For each dataset, we provide the distribution of all zero-shot relations' frequncies in Fig. \ref{fig: data comparison}. We take the relations with lowest frequencies as zero-shot relations when we construct datasets, following previous few-shot relational TKG learning frameworks, e.g., OAT \cite{DBLP:conf/akbc/MirtaheriR0MG21} and MOST \cite{DBLP:conf/ijcnn/DingHWMHT23}. 
The proportion of zero-shot relations for each dataset is high. 14 out of 23; 123 out of 253; 155 out of 248 relations in ACLED-zero; ICEWS21-zero; ICEWS22-zero are zero-shot relations. This ensures the diversity of relation types in test sets.

\section{Implementation Details}
\label{app: implementation details}
All experiments are implemented with PyTorch \cite{DBLP:conf/nips/PaszkeGMLBCKLGA19} on a server equipped with an AMD EPYC 7513 32-Core Processor and a single NVIDIA A40 with 48GB memory. All the experimental results are the average of three runs with different random seeds. 
\subsection{Baseline Implementation Details}
Our baselines are all based on neural networks rather than pure score function-based (e.g., TTransE \cite{DBLP:conf/www/LeblayC18}). This is because the most popular and recent TKGF methods all leverage neural networks to gain the forecasting ability and it is hard for pure score function-based methods to achieve that solely with geometric embeddings. The implementation details of each TKGF baseline is as follows.
\begin{itemize}
    \item \textbf{CyGNet.} We use the official code of CyGNet\footnote{https://github.com/CunchaoZ/CyGNet}. We search hyperparameters of baseline CyGNet following Table \ref{tab: cygnet hyperparameter search}. The best hyperparameters are marked as bold. For each dataset, we do 4 trials to try different hyperparameter settings. We run 5 epochs for each trail and take the one with the best validation result as the best hyperparameter setting. 
    \begin{table}[htbp]
      \centering
      \resizebox{\columnwidth}{!}{
        \begin{tabular}{lccc}
            \toprule 
       \textbf{Dataset} & \textbf{ACLED-zero} & \textbf{ICEWS21-zero} & \textbf{ICEWS22-zero}\\
       \textbf{Hyperparameter} & CyGNet & CyGNet & CyGNet\\
       \midrule 
       Embedding Size & \{100, \textbf{200}\} & \{100, \textbf{200}\} & \{100, \textbf{200}\} \\
       Alpha (Eq. 9 in \cite{DBLP:conf/aaai/ZhuCFCZ21})&  \{\textbf{0.2}, 0.5\} &  \{\textbf{0.2}, 0.5\} &  \{\textbf{0.2}, 0.5\}  \\
       
      \bottomrule 
        \end{tabular}
        }
    \caption{CyGNet hyperparameter searching strategy.}
    \label{tab: cygnet hyperparameter search}
    \end{table}
    \item \textbf{TANGO-TuckER/Distmult.} We use the official code of TANGO\footnote{https://github.com/TemporalKGTeam/TANGO}. We search hyperparameters of baseline TANGO-TuckER/Distmult following Table \ref{tab: tango hyperparameter search}. The best hyperparameters are marked as bold. For each dataset, we do 6 (TANGO-TuckER) and 9 (TANGO-Distmult) trials to try different hyperparameter settings. We run 10 epochs for each trail and take the one with the best validation result as the best hyperparameter setting. 
    \begin{table}[htbp]
      \centering
      \resizebox{\columnwidth}{!}{
        \begin{tabular}{lcccccc}
            \toprule 
       \textbf{Dataset} & \multicolumn{2}{c}{\textbf{ACLED-zero}} & \multicolumn{2}{c}{\textbf{ICEWS21-zero}} & \multicolumn{2}{c}{\textbf{ICEWS22-zero}}\\
       \cmidrule(lr){2-3} \cmidrule(lr){4-5} \cmidrule(lr){6-7}
       \textbf{Hyperparameter} & TuckER & Distmult & TuckER & Distmult & TuckER & Distmult\\
       \midrule 
       Embedding Size & \{100, \textbf{200}\} & \{100, 200, \textbf{300}\} & \{\textbf{100}, 200\} & \{\textbf{100}, 200, 300\} & \{100, \textbf{200}\} & \{100, \textbf{200}, 300\} \\
       History Length &  \{\textbf{4}, 6, 10\} &  \{\textbf{4}, 6, 10\} &  \{\textbf{4}, 6, 10\} &  \{\textbf{4}, 6, 10\} &  \{\textbf{4}, 6, 10\} &  \{\textbf{4}, 6, 10\} \\
       
      \bottomrule 
        \end{tabular}
        }
    \caption{TANGO hyperparameter searching strategy.}
    \label{tab: tango hyperparameter search}
    \end{table}
    \item \textbf{RE-GCN.} We use the official code of RE-GCN\footnote{https://github.com/Lee-zix/RE-GCN}. We search hyperparameters of baseline RE-GCN following Table \ref{tab: RE-GCN hyperparameter search}. The best hyperparameters are marked as bold. For each dataset, we do 4 trials to try different hyperparameter settings. We run 10 epochs for each trail and take the one with the best validation result as the best hyperparameter setting. 
    \begin{table}[htbp]
      \centering
      \resizebox{0.8\columnwidth}{!}{
        \begin{tabular}{lccc}
            \toprule 
       \textbf{Dataset} & \textbf{ACLED-zero} & \textbf{ICEWS21-zero} & \textbf{ICEWS22-zero}\\
       \textbf{Hyperparameter} & RE-GCN & RE-GCN & RE-GCN\\
       \midrule 
       Embedding Size & \{100, \textbf{200}\} & \{\textbf{100}, 200\} & \{100, \textbf{200}\} \\
       History Length &  \{\textbf{3}, 9\} &  \{3, \textbf{9}\} &  \{3, \textbf{9}\}  \\
       
      \bottomrule 
        \end{tabular}
        }
    \caption{RE-GCN hyperparameter searching strategy.}
    \label{tab: RE-GCN hyperparameter search}
    \end{table}
    \item \textbf{TiRGN.} We use the official code of TiRGN\footnote{https://github.com/Liyyy2122/TiRGN}. We search hyperparameters of baseline TiRGN following Table \ref{tab: TiRGN hyperparameter search}. The best hyperparameters are marked as bold. For each dataset, we do 12 trials to try different hyperparameter settings. We run 10 epochs for each trail and take the one with the best validation result as the best hyperparameter setting.
        \begin{table}[htbp]
      \centering
      \resizebox{\columnwidth}{!}{
        \begin{tabular}{lccc}
            \toprule 
       \textbf{Dataset} & \textbf{ACLED-zero} & \textbf{ICEWS21-zero} & \textbf{ICEWS22-zero}\\
       \textbf{Hyperparameter} & TiRGN & TiRGN & TiRGN\\
       \midrule 
       Embedding Size & \{100, \textbf{200}\} & \{\textbf{100}, 200\} & \{100, \textbf{200}\} \\
       History Length &  \{\textbf{3}, 9\} &  \{3, \textbf{9}\} &  \{3, \textbf{9}\}  \\
       Alpha (Eq. 11 in \cite{DBLP:conf/ijcai/LiS022})  &  \{\textbf{0.3}, 0.5, 0.7\} &  \{\textbf{0.3}, 0.5, 0.7\} &  \{\textbf{0.3}, 0.5, 0.7\}  \\
       
      \bottomrule 
        \end{tabular}
        }
    \caption{TiRGN hyperparameter searching strategy.}
    \label{tab: TiRGN hyperparameter search}
    \end{table}
    \item \textbf{RETIA.} We use the official code of RETIA\footnote{https://github.com/CGCL-codes/RETIA}. We search hyperparameters of baseline RETIA following Table \ref{tab: RETIA hyperparameter search}. The best hyperparameters are marked as bold. For each dataset, we do 4 trials to try different hyperparameter settings. We run 10 epochs for each trail and take the one with the best validation result as the best hyperparameter setting.
        \begin{table}[htbp]
      \centering
      \resizebox{0.8\columnwidth}{!}{
        \begin{tabular}{lccc}
            \toprule 
       \textbf{Dataset} & \textbf{ACLED-zero} & \textbf{ICEWS21-zero} & \textbf{ICEWS22-zero}\\
       \textbf{Hyperparameter} & RETIA & RETIA & RETIA\\
       \midrule 
       Embedding Size & \{100, \textbf{200}\} & \{\textbf{100}, 200\} & \{100, \textbf{200}\} \\
       History Length &  \{\textbf{3}, 9\} &  \{3, \textbf{9}\} &  \{3, \textbf{9}\}  \\
       
      \bottomrule 
        \end{tabular}
        }
    \caption{RETIA hyperparameter searching strategy.}
    \label{tab: RETIA hyperparameter search}
    \end{table}
    \item \textbf{CENET.} We use the official code of CENET\footnote{https://github.com/xyjigsaw/CENET}. We search hyperparameters of baseline CENET following Table \ref{tab: CENET hyperparameter search}. The best hyperparameters are marked as bold. For each dataset, we do 4 trials to try different hyperparameter settings. We run 5 epochs for each trail and take the one with the best validation result as the best hyperparameter setting.
    \begin{table}[htbp]
      \centering
      \resizebox{0.8\columnwidth}{!}{
        \begin{tabular}{lccc}
            \toprule 
       \textbf{Dataset} & \textbf{ACLED-zero} & \textbf{ICEWS21-zero} & \textbf{ICEWS22-zero}\\
       \textbf{Hyperparameter} & CENET & CENET & CENET\\
       \midrule 
       Embedding Size & \{\textbf{100}, 200\} & \{100, \textbf{200}\} & \{100, \textbf{200}\} \\
       Mask Strategy &  \{\textbf{soft}, hard\} &  \{\textbf{soft}, hard\} &  \{\textbf{soft}, hard\}  \\
       
      \bottomrule 
        \end{tabular}
        }
    \caption{CENET hyperparameter searching strategy.}
    \label{tab: CENET hyperparameter search}
    \end{table}
\end{itemize}
The hyperparameters not discussed above follow the settings reported in the original papers. 
\begin{table*}[htbp]
      \centering
      \resizebox{\textwidth}{!}{
        \begin{tabular}{lcccccccccccc}
            \toprule 
       \textbf{Dataset} & \multicolumn{4}{c}{\textbf{ACLED-zero}} & \multicolumn{4}{c}{\textbf{ICEWS21-zero}} & \multicolumn{4}{c}{\textbf{ICEWS22-zero}}\\
       \cmidrule(lr){2-5} \cmidrule(lr){6-9} \cmidrule(lr){10-13}
       \textbf{Model} & $\alpha$ & $\gamma$ Type & $\gamma$ Value & $\eta$ & $\alpha$ & $\gamma$ Type & $\gamma$ Value & $\eta$ & $\alpha$ & $\gamma$ Type & $\gamma$ Value & $\eta$\\
       \midrule 
       CyGNet+ & \{1, \textbf{0.1}\} & \{\textbf{Fixed}, Unfixed\} & \{\textbf{1}, 0.01, 0.001\} & \{1.2, \textbf{1}\} 
               & \{1, \textbf{0.1}\} & \{\textbf{Fixed}, Unfixed\} & \{1, 0.01, \textbf{0.001}\} & \{1.2, \textbf{1}\}
               & \{1, \textbf{0.1}\} & \{\textbf{Fixed}, Unfixed\} & \{1, 0.01, \textbf{0.001}\} & \{1.2, \textbf{1}\}
       \\
       TANGO-T+ & \{1, \textbf{0.1}\} & \{Fixed, \textbf{Unfixed}\} & \{\textbf{1}, 0.01, 0.001\} & \{\textbf{1.2}, 1\} 
               & \{1, \textbf{0.1}\} & \{Fixed, \textbf{Unfixed}\} & \{\textbf{1}, 0.01, 0.001\} & \{\textbf{1.2}, 1\}
               & \{1, \textbf{0.1}\} & \{\textbf{Fixed}, Unfixed\} & \{\textbf{1}, 0.01, 0.001\} & \{\textbf{1.2}, 1\}
        \\
       TANGO-D+& \{1, \textbf{0.1}\} & \{\textbf{Fixed}, Unfixed\} & \{\textbf{1}, 0.01, 0.001\} & \{\textbf{1.2}, 1\} 
               & \{1, \textbf{0.1}\} & \{Fixed, \textbf{Unfixed}\} & \{\textbf{1}, 0.01, 0.001\} & \{\textbf{1.2}, 1\}
               & \{1, \textbf{0.1}\} & \{Fixed, \textbf{Unfixed}\} & \{\textbf{1}, 0.01, 0.001\} & \{\textbf{1.2}, 1\}
        \\
        RE-GCN+& \{1, \textbf{0.1}\} & \{\textbf{Fixed}, Unfixed\} & \{1, 0.01, \textbf{0.001}\} & \{1.2, \textbf{1}\} 
               & \{1, \textbf{0.1}\} & \{\textbf{Fixed}, Unfixed\} & \{1, \textbf{0.01}, 0.001\} & \{1.2, \textbf{1}\}
               & \{1, \textbf{0.1}\} & \{\textbf{Fixed}, Unfixed\} & \{1, \textbf{0.01}, 0.001\} & \{1.2, \textbf{1}\}
        \\
        TiRGN+& \{1, \textbf{0.1}\} & \{Fixed, \textbf{Unfixed}\} & \{1, 0.01, \textbf{0.001}\} & \{1.2, \textbf{1}\} 
               & \{1, \textbf{0.1}\} & \{\textbf{Fixed}, Unfixed\} & \{1, \textbf{0.01}, 0.001\} & \{1.2, \textbf{1}\}
               & \{1, \textbf{0.1}\} & \{\textbf{Fixed}, Unfixed\} & \{1, \textbf{0.01}, 0.001\} & \{1.2, \textbf{1}\}
        \\
        RETIA+& \{1, \textbf{0.1}\} & \{Fixed, \textbf{Unfixed}\} & \{1, \textbf{0.01}, 0.001\} & \{\textbf{2}, 1\} 
               & - & - & - & -
               & \{1, \textbf{0.1}\} & \{\textbf{Fixed}, Unfixed\} & \{1, \textbf{0.01}, 0.001\} & \{\textbf{2}, 1\}
        \\
        CENET+& \{1, \textbf{0.1}\} & \{Fixed, \textbf{Unfixed}\} & \{\textbf{1}, 0.01, 0.001\} & \{1.2, \textbf{1}\} 
               & \{1, \textbf{0.1}\} & \{Fixed, \textbf{Unfixed}\} & \{\textbf{1}, 0.01, 0.001\} & \{1.2, \textbf{1}\}
               & \{1, \textbf{0.1}\} & \{Fixed, \textbf{Unfixed}\} & \{1, \textbf{0.01}, 0.001\} & \{1.2, \textbf{1}\}
        \\
       
      \bottomrule 
        \end{tabular}
        }
    \caption{zrLLM hyperparameter searching strategy. The best settings are marked as bold.}
    \label{tab: zrLLM hyperparameter search}
    \end{table*}
    \begin{table*}[htbp]
      \centering
      \resizebox{0.83\textwidth}{!}{
        \begin{tabular}{lcccccc}
            \toprule 
       \textbf{Dataset} & \multicolumn{2}{c}{\textbf{ACLED-zero}} & \multicolumn{2}{c}{\textbf{ICEWS21-zero}} & \multicolumn{2}{c}{\textbf{ICEWS22-zero}}\\
       \cmidrule(lr){2-3} \cmidrule(lr){4-5} \cmidrule(lr){6-7}
       \textbf{Model} & Training Time (h) & GPU Memory (MB) & Training Time (h) & GPU Memory (MB) & Training Time (h) & GPU Memory (MB)\\
       \midrule 
       CyGNet+ & 0.03 & 2,216 & 17.87 & 7,470 & 4.80 & 9,574
       \\
       TANGO-T+ & 0.05 & 2,716 & 8.64 & 34,186 & 2.82 & 20,120
        \\
       TANGO-D+ & 0.11 & 3,064 & 10.88 & 34,034 & 0.70 & 19,250
        \\
        RE-GCN+ & 0.06 & 1,587 & 14.70 & 26,420 & 3.85 & 19,168
        \\
        TiRGN+ & 0.10 & 2,654 & 11.67 & 36,780 & 2.40 & 15,976
        \\
        RETIA+ & 0.13 & 4,274 & - & - & 9.33 & 26,328
        \\
        CENET+ & 0.03 & 1,429 & 48.94 & 6,750 & 12.54 & 5,639
        \\
        PPT & 0.47 & 7,654 & 84.68 & 9,078 & 59.35 & 7,678
        \\
       
      \bottomrule 
        \end{tabular}
        }
    \caption{Computational resources required by zrLLM-enhanced models and PPT.}
    \label{tab: Computational Resource}
    \end{table*}
\subsection{zrLLM Implementation Details}
We fix the hyperparameters searched from the baselines and additionally search zrLLM-specific hyperparameters for zrLLM-enhanced models.
The hyperparameter searching strategy and the best hyperparameter settings regarding the zrLLM-enhanced baselines are reported in Table \ref{tab: zrLLM hyperparameter search}. Note that $\gamma$ can be either a learnable parameter or a fixed scalar. When $\gamma$ is not fixed, $\gamma$ Value means the initialized parameter value during training. For each zrLLM-enhanced model, in each dataset, we do 36 trials to try different hyperparameter settings. We run 7 epochs for each trail and take the one with the best validation result as the best hyperparameter setting.
\subsection{Implementation Details of PPT and ICL}
\label{app: implementation details PPT icl}
We use the official code of PPT\footnote{https://github.com/JaySaligia/PPT} and ICL\footnote{https://github.com/usc-isi-i2/isi-tkg-icl}. For PPT, we use the default hyperparameter setting used for ICEWS14 when we implement it on all our new datasets. Since PPT only explores object entity prediction in its original implementation, we add the subject entity prediction part and report the overall result. We achieve subject prediction by first deriving the inverse relation texts for each relation in each TKG dataset, e.g., use \textit{Inversed Reduce or stop military assistance} to represent the inverse relation of the relation \textit{Reduce or stop military assistance}, and then turning each subject prediction query $(?,r_q,o_q, t_q)$ to an object prediction query $(o_q,r^{-1}_q,?, t_q)$, where $r^{-1}_q$ stands for the inverse relation of $r_q$. For ICL, we use the lexical-based prompt because we are dealing with zero-shot relations where text information is important. We also employ the unidirectional entity-focused history, which achieves best results on ICEWS14 as reported in ICL's original paper. We use the default history length of 20 for all datasets.

\subsection{Computational Resource Usage}
\label{app: space and time}
\begin{figure*}[htbp]
    \centering
  \subfloat[ACLED-zero.\label{fig: comp acled}]{%
       \includegraphics[width=0.33\textwidth]{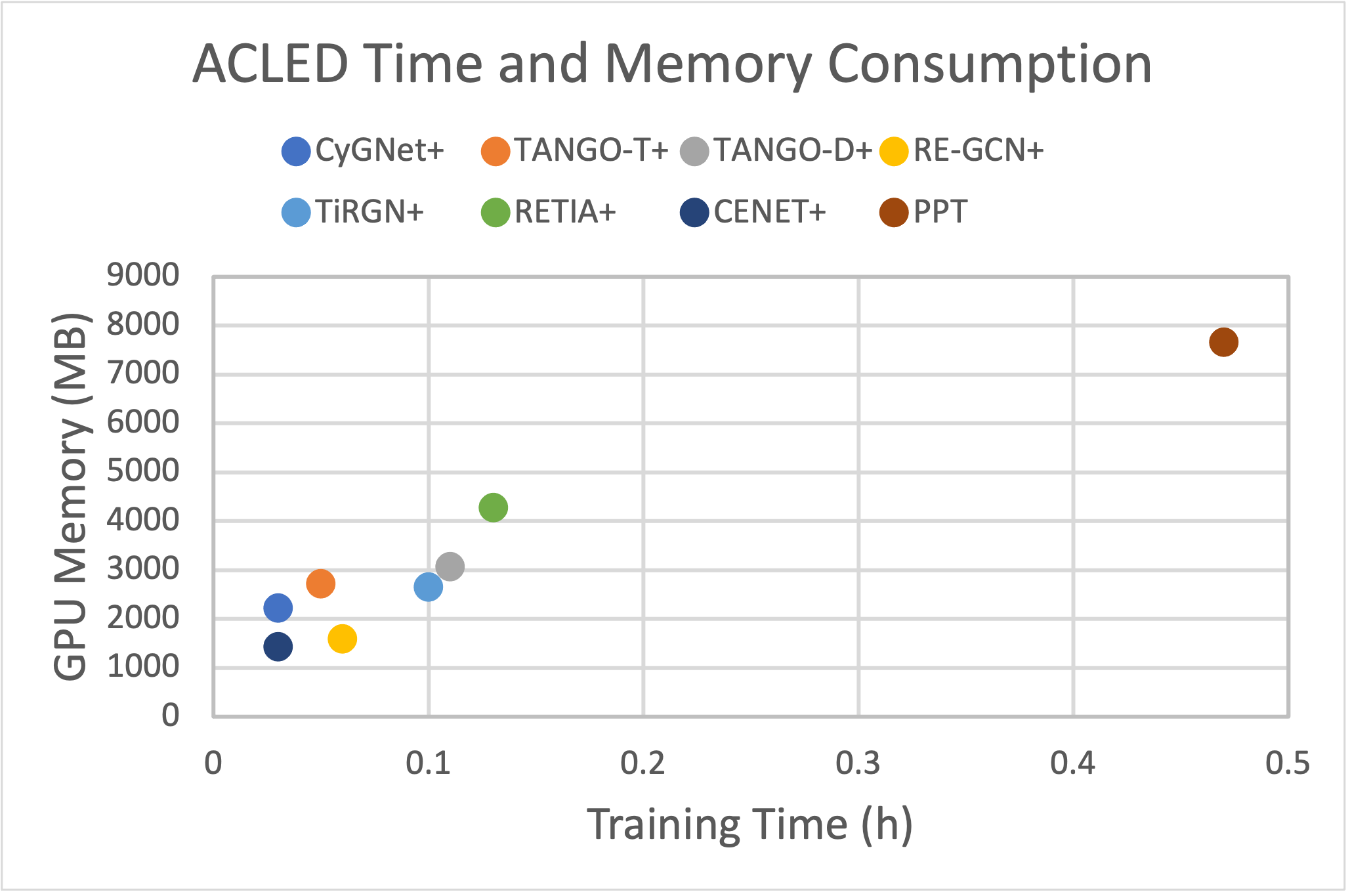}}
    \hfill
  \subfloat[ICEWS21-zero.\label{fig: comp 21}]{%
        \includegraphics[width=0.33\textwidth]{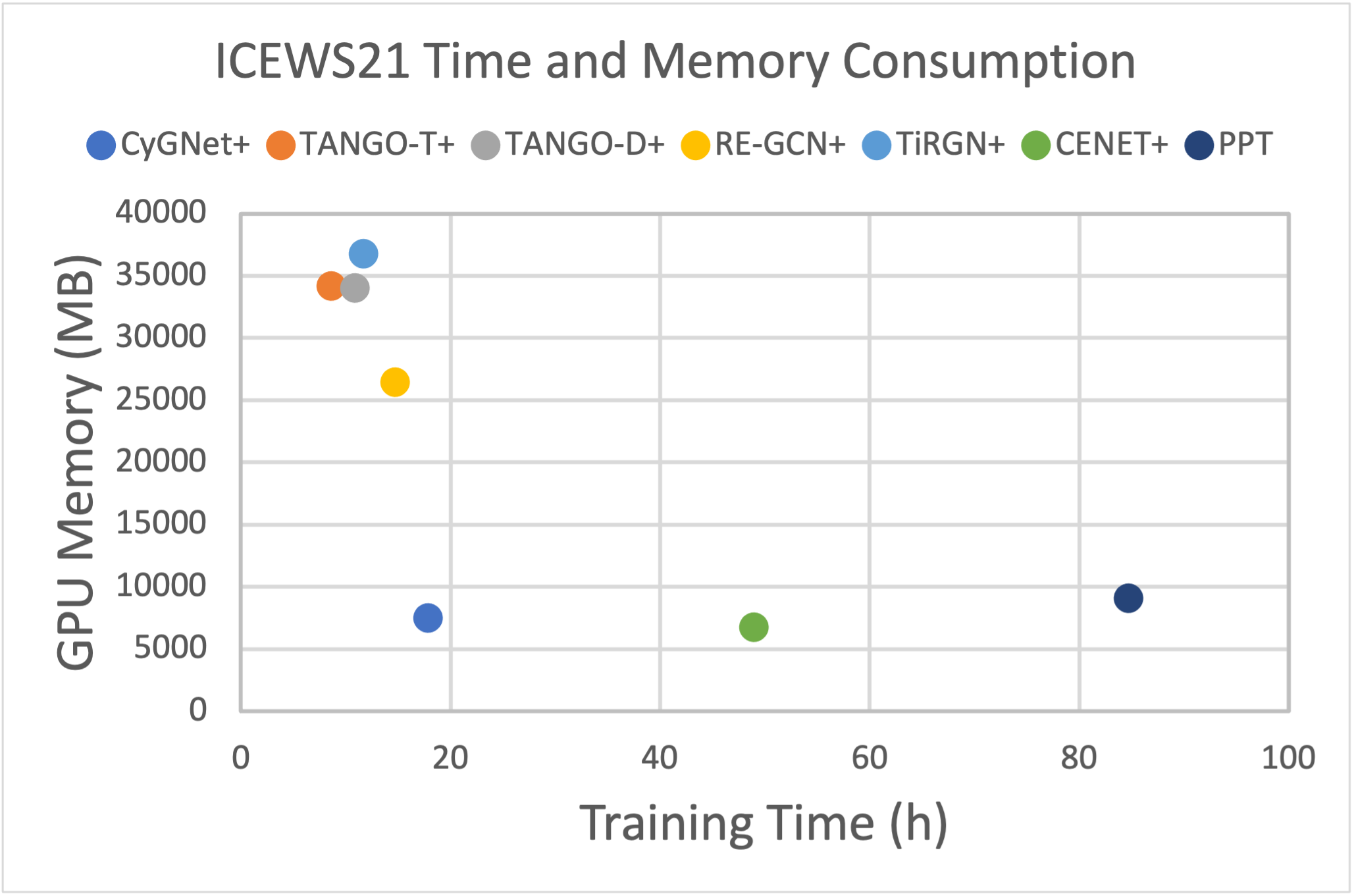}}
    \hfill
  \subfloat[ICEWS22-zero.\label{fig: comp 22}]{%
        \includegraphics[width=0.33\textwidth]{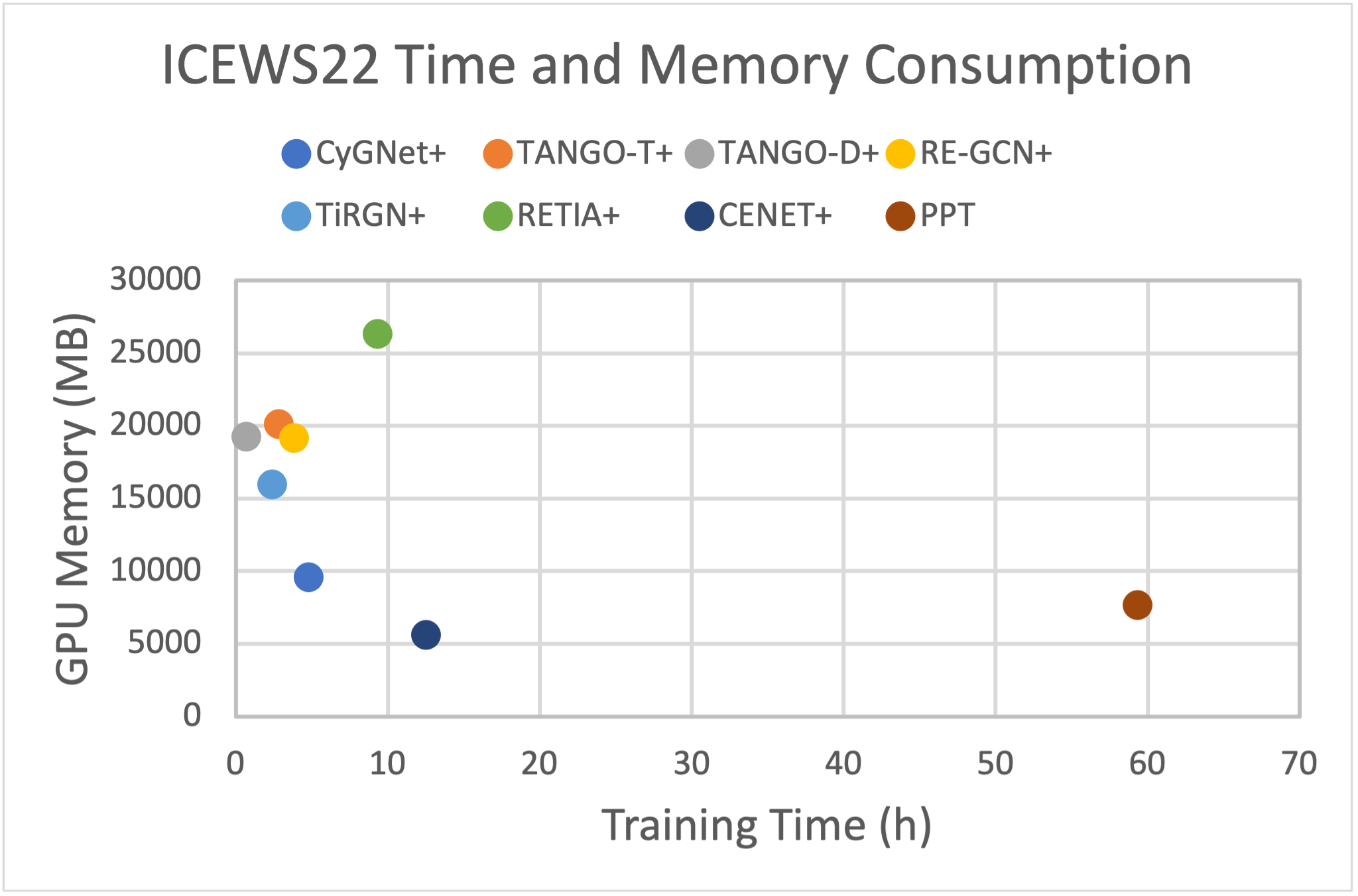}}
    \hfill
  \caption{Computational resources required during training of zrLLM-enhanced models and PPT.}
  \label{fig: comp} 
\end{figure*}

We report the computational resources for all zrLLM-enhanced models and PPT in Table \ref{tab: Computational Resource}. Training time denotes the period of time a model requires to reach its best validation performance. 
PPT requires extremely long time for sampling and thus has high time consumption. 
Note that zrLLM loads T5 to generate LM-based relation representations. This process takes a substantial amount of GPU memory. However, in our work, we store the output of T5's encoder as saved parameters and use them in downstream zero-shot TKGF with any zrLLM-enhanced model. This prevents from high memory demand during model training and evaluation. We use Fig. \ref{fig: comp} to illustrate the direct comparison among zrLLM-enhanced models and PPT regarding their required computational resources during training. 

ICL loads GPT-NeoX-20B that requires huge memory consumption. We use two NVIDIA A40 for all its experiments. Since ICL does not require training, we only report its validation and test time here. For ACLED-zero, GPU memory usage is 90,846 MB. Validation time is 0.63 h and test time is 0.12 h. For ICEWS21-zero, GPU memory usage is 90,868 MB. Validation time is 35.48 h and test time is 0.82 h. For ICEWS22-zero, GPU memory usage is 91,458 MB. Validation time is 22.98 h and test time is 1.15 h.
\subsection{Zero-Shot Evaluation Setting Explanation}
\label{app: zero setting}
To keep zero-shot relations "always unseen" during the whole evaluation process, we constrain all models to do LP only based on the training set. 
Among all TKGF models, TANGO, RE-GCN, TiRGN and RETIA use recurrent neural structures to model historical TKG information from a short sequence of timestamps prior to the prediction timestamp. We constrain them to only use the latest training data, i.e., from $t_\text{train\_max} - k$ to $t_\text{train\_max}$, to encode historical information during evaluation. $k$ is the considered history length and $t_\text{train\_max} = \text{max}(\mathcal{T}_\text{train})$ is the maximum timestamp in the training data. For CyGNet and CENET, they have originally met our restriction of not observing any ground truth evaluation data during evaluation, and thus can be directly implemented in our zero-shot setting. Another point worth noting is that RHL requires ground truth relation history. We restrict zrLLM to only capture the relation history across the whole training time period to prevent from exposing zero-shot relations during evaluation.
\section{Algorithm}
\label{app: alg}
We provide algorithms to show the whole process of using zrLLM to enhance TKGF models. First, zrLLM generates LLM-based relation representations by using GPT-3.5 and T5-11B (Algorithm \ref{alg: generate text emb}). Then we train zrLLM jointly with TKGF baseline models (Algorithm \ref{alg: alg1}). The trained models are then used for evaluation (Algorithm \ref{alg: alg2}).
\begin{algorithm}[htbp]
\scriptsize
\caption{Generate LLM-based Relation Representations}
\label{alg: generate text emb}

\DontPrintSemicolon
\KwInput{Relations $\mathcal{R}$, relation text of all relations provided by the TKG dataset $\text{T}\textsc{ext}_\mathcal{R}$}

\For{\textup{batch =} 1: B}
{
Take a batch of $n$ relations from $\mathcal{R}$

Pick out their relation texts from $\text{T}\textsc{ext}_\mathcal{R}$

Write prompt with the relation texts \tcp*[h]{Fig. \ref{fig: prompt}}

Input the prompt into GPT-3.5

Extract the ERDs from the output of GPT-3.5

Input the ERDs into T5-11B's encoder

Store the output of T5-11B's encoder
}

return T5-encoded text representation $\Bar{\mathbf{H}}_r$ for every $r \in \mathcal{R}$

\end{algorithm}
\begin{algorithm}[htbp]
\scriptsize
\caption{Model Training with zrLLM}
\label{alg: alg1}

\DontPrintSemicolon
\KwInput{Entities $\mathcal{E}$, relations $\mathcal{R}$, timestamps $\mathcal{T}$, T5-encoded text representations $\{\Bar{\mathbf{H}}_r\}$ for $\mathcal{R}$, training set $\mathcal{G}_{\text{train}}$}
Align $\{\Bar{\mathbf{H}}_r\}$ to TKG embedding space and get $\{\Bar{\mathbf{h}}_r\}$ \tcp*[h]{Eq. \ref{eq: align1}, \ref{eq: align2}}

\For{\textup{epoch =} 1: V}
{

\For{\textup{batch =} 1: B}
{

Take a batch of training facts $\{(s,r,o,t)\} \in \mathcal{G}_{\text{train}}$

Find the relation history of $s$ and $o$ before $t$ for each $(s,r,o,t)$

Encode relation history until $t-1$ \tcp*[h]{Eq. \ref{eq: hist}}

Compute the predicted history with HPN \tcp*[h]{Eq. \ref{eq: path pred}}

Compute history-related MSE loss $\mathcal{L}_\text{hist}$ \tcp*[h]{Eq. \ref{eq: mse}}

Compute the representation of the $r$-related temporal relation pattern \tcp*[h]{Eq. \ref{eq: final path}}

Compute the RHL-based score \tcp*[h]{Eq. \ref{eq: rhl score}}

Input $\{\Bar{\mathbf{h}}_r\}$ into TKGF baseline and compute LP score

Compute total score for the training batch \tcp*[h]{Eq. \ref{eq: total score}}

Compute TKGF model loss $\mathcal{L}_\text{TKGF}$ \tcp*[h]{Eq. \ref{eq: tkgf loss}}

Compute RHL-based loss $\mathcal{L}_\text{RHL}$

Compute total loss $\mathcal{L}_\text{total}$ \tcp*[h]{Eq. \ref{eq: total loss}}

Update model parameters using gradient of $\bigtriangledown	\mathcal{L}_\text{total}$
}
}
return trained zrLLM-enhanced TKGF model

\end{algorithm}
\begin{algorithm}[htbp]
\scriptsize
\caption{Model Evaluation with zrLLM}
\label{alg: alg2}

\DontPrintSemicolon
\KwInput{Entities $\mathcal{E}$, relations $\mathcal{R}$, timestamps $\mathcal{T}$, LLM-based relation representations $\{\Bar{\mathbf{h}}_r\}$ for $\mathcal{R}$, training set $\mathcal{G}_{\text{train}}$, validation set $\mathcal{G}_{\text{valid}}$, test set $\mathcal{G}_{\text{test}}$}
\uIf{evaluation set is $\mathcal{G}_{\text{valid}}$}{
    $\mathcal{G}_{\text{eval}} = \mathcal{G}_{\text{valid}}$
  }
\Else{
    $\mathcal{G}_{\text{eval}} = \mathcal{G}_{\text{test}}$
 }
\For{\textup{batch =} 1: B}
{

Take a batch of evaluation facts $\{(s_q,r_q,o_q,t_q)\} \in \mathcal{G}_{\text{eval}}$

Derive LP queries $\{(s_q,r_q,?,t_q)\}$

Input $\{r_q\}$ into HPN and compute the predicted history \tcp*[h]{Eq. \ref{eq: path pred}}

Compute the representation of the $r_q$-related temporal relation pattern for each LP query \tcp*[h]{Eq. \ref{eq: final path}}

Compute the RHL-based score of each candidate entity $e\in \mathcal{E}$ for each LP query \tcp*[h]{Eq. \ref{eq: rhl score}}

Input $\{\Bar{\mathbf{h}}_r\}$ into TKGF baseline and compute LP score of each candidate entity $e\in \mathcal{E}$ for each LP query

Compute total score of each candidate entity $e\in \mathcal{E}$ for each LP query in the batch \tcp*[h]{Eq. \ref{eq: total score}}

Rank candidate entities $\mathcal{E}$ with their total scores in the descending order

Compute and record the rank of the ground truth missing entity $o_q$ for each LP query
}

Compute MRR and Hits@1/3/10

return MRR and Hits@1/3/10

\end{algorithm}

\section{Evaluation Metrics Details}
\label{app: metric}
We employ two evaluation metrics, i.e., mean reciprocal
rank (MRR) and Hits@1/3/10. For every LP query $q$, we compute the rank $\theta_q$ of the ground truth missing entity. We define MRR as: $\frac{1}{|\mathcal{G}_{\text{test}}|}\sum_{q} \frac{1}{\theta_q}$ (the definition is similar for $\mathcal{G}_{\text{valid}}$). Hits@1/3/10 denote the proportions of the predicted links where ground truth missing entities are ranked as top 1, top3, top10, respectively. As explored and suggested in \cite{DBLP:conf/pkdd/GastingerSSSS23}, we also use the time-aware filtering setting proposed in \cite{DBLP:conf/iclr/HanCMT21} for fairer evaluation.
\section{Complete Comparative Study Results}
\label{app: complete main results}
We report the complete results of comparative study in Table \ref{tab: tid LP results first two dataset} and \ref{tab: tid LP results}.
\begin{table*}[htbp]
    \centering
    \resizebox{\textwidth}{!}{
    \large\begin{tabular}{@{}lcccc cccc ccccc ccccc@{}}
\toprule
        \textbf{Datasets} & \multicolumn{9}{c}{\textbf{ICEWS21-zero}} & \multicolumn{9}{c}{\textbf{ICEWS22-zero}} \\
\cmidrule(lr){2-10} \cmidrule(lr){11-19}
        & \multicolumn{4}{c}{Zero-Shot Relations} & \multicolumn{4}{c}{Seen Relations} & Overall & \multicolumn{4}{c}{Zero-Shot Relations} & \multicolumn{4}{c}{Seen Relations} & Overall\\
\cmidrule(lr){2-5} \cmidrule(lr){6-9} \cmidrule(lr){10-10} \cmidrule(lr){11-14} \cmidrule(lr){15-18} \cmidrule(lr){19-19}
        \textbf{Model} & MRR & Hits@1 & Hits@3 & Hits@10 & MRR & Hits@1 & Hits@3 & Hits@10 & MRR  & MRR & Hits@1 & Hits@3 & Hits@10 & MRR & Hits@1 & Hits@3 & Hits@10 & MRR \\
\midrule 
        CyGNet 
        & 0.120 & 0.046 & 0.130 & 0.270
        & 0.254 & \textbf{0.165} & 0.293 & 0.432
        & 0.252
        & 0.211	& 0.098	& 0.240	& 0.459
        & \textbf{0.315}	& 0.198	& \textbf{0.373}	& 0.540
        & 0.311
         \\
        CyGNet+
        & \textbf{0.201}	& \textbf{0.103}	& \textbf{0.226}	& \textbf{0.415} 
        & \textbf{0.258}	& 0.162	& \textbf{0.294}	& \textbf{0.447} & \textbf{0.257}
        & \textbf{0.286}	& \textbf{0.167}	& \textbf{0.324}	& \textbf{0.542}
        & \textbf{0.315}	& \textbf{0.200}	& 0.364	& \textbf{0.545} & \textbf{0.314}
        \\
\midrule
        TANGO-T 
        & 0.067 & 0.031 & 0.069	& 0.132
        & \textbf{0.283}	& \textbf{0.190}	& \textbf{0.319}	& \textbf{0.470}
        & \textbf{0.279}
        & 0.092 & 0.042 & 0.100	& 0.187
        & \textbf{0.363}	& 0.250	& 0.407	& 0.579
        & 0.352

        \\
        TANGO-T+ 
        & \textbf{0.216}	& \textbf{0.125} & \textbf{0.245} & \textbf{0.395} 
        & 0.280	& 0.186 & 0.313 & 0.466 & \textbf{0.279}
        & \textbf{0.326}	& \textbf{0.198} & \textbf{0.388} & \textbf{0.578} 
        & \textbf{0.363}	& \textbf{0.251} & \textbf{0.409} & \textbf{0.585} & \textbf{0.362}
        \\
\midrule       
       TANGO-D
       & 0.012		& 0.005 & 0.011		& 0.023
       & 0.266	& \textbf{0.178} & 0.298 & 0.439
       & 0.261
       & 0.011		& 0.002 & 0.007		& 0.018
       & \textbf{0.350}	& 0.227 & \textbf{0.394} & 0.569
       & 0.337

        \\
       TANGO-D+
       & \textbf{0.212}	& \textbf{0.122} & \textbf{0.237}		& \textbf{0.400} 
       & \textbf{0.268}	& 0.175 & \textbf{0.303} & \textbf{0.453}   & \textbf{0.267}
       & \textbf{0.311}	& \textbf{0.186}	& \textbf{0.374}	& \textbf{0.574}	
        & \textbf{0.350}	& \textbf{0.239}	& 0.393	& \textbf{0.570} & \textbf{0.348}
        \\
\midrule
        RE-GCN
        & 0.200	& 0.104 & 0.231	& 0.379
        & 0.277	& 0.185 & 0.309 & \textbf{0.456} 
        & 0.276
        & 0.280	& 0.162 & 0.321	& \textbf{0.616}
        & 0.354	& 0.243 & \textbf{0.398} & 0.567
        & 0.351
        \\
        RE-GCN+ 
        & \textbf{0.214}	& \textbf{0.117} & \textbf{0.246} & \textbf{0.406} 
        & \textbf{0.280}	& \textbf{0.188} & \textbf{0.314} & \textbf{0.456} & \textbf{0.279}
        & \textbf{0.324}	& \textbf{0.194} & \textbf{0.376} & 0.595
        & \textbf{0.357}	& \textbf{0.244} & \textbf{0.398} & \textbf{0.573} & \textbf{0.356}

        \\
\midrule
        TiRGN
        & 0.189	& 0.101	& 0.209	& 0.368
        & 0.275	& 0.182	& 0.308	& 0.457
        & 0.273
        & 0.299	& 0.169	& 0.358	& 0.570
        & 0.352	& 0.239	& 0.399	& 0.575
        & 0.350
        \\
        TiRGN+ 
        & \textbf{0.221}	& \textbf{0.130}	& \textbf{0.246}	& \textbf{0.410}	
        & \textbf{0.279}	& \textbf{0.185}	& \textbf{0.323}	& \textbf{0.464} & \textbf{0.278}
        & \textbf{0.333}	& \textbf{0.203}	& \textbf{0.383}	& \textbf{0.602}	
        & \textbf{0.353}	& \textbf{0.240}	& \textbf{0.400}	& \textbf{0.577} & \textbf{0.352}
        \\
\midrule
RETIA 
        &&&
        \multicolumn{5}{c}{\multirow{2}{*}{\text{>> 120 Hours Timeout}}}
        &&
        & 0.302	& 0.166	& 0.349	& 0.566	
        & 0.356	& 0.245	& 0.401	& 0.577
        & 0.354
        \\
        RETIA+ 
        & &&&&&&&&
        & \textbf{0.331}	& \textbf{0.201}	& \textbf{0.384}	& \textbf{0.597}	
        & \textbf{0.358}	& \textbf{0.247}	& \textbf{0.402}	& \textbf{0.578} & \textbf{0.357}
        \\
\midrule
        CENET
        & 0.205	& 0.101	& 0.232	& 0.411	
        & 0.288	& 0.196	& 0.318	& 0.468
        & 0.287
        & 0.270	& 0.134	& 0.318	& 0.544	
        & 0.379	& 0.268	& 0.423	& 0.599
        & 0.375
        \\
        CENET+ 
        & \textbf{0.335}	& \textbf{0.162}	& \textbf{0.455}	& \textbf{0.659}		
        & \textbf{0.396}	& \textbf{0.239}	& \textbf{0.502}	& \textbf{0.688} & \textbf{0.395}
        & \textbf{0.564}	& \textbf{0.432}	& \textbf{0.649}	& \textbf{0.801}	
        & \textbf{0.571}	& \textbf{0.451}	& \textbf{0.651}	& \textbf{0.773} & \textbf{0.571}
        \\
\midrule
        PPT
        & 0.212 & 0.120 & 0.240 & 0.403
        & 0.269 & 0.172 & 0.304 & 0.462
        & 0.268
        & 0.323 & 0.191 & 0.376 & 0.598
        & 0.332 & 0.219 & 0.377 & 0.556
        & 0.331
        \\
\midrule 
        ICL
        & 0.156 & 0.096 & 0.180 & 0.300
        & 0.178 & 0.120 & 0.206 & 0.308
        & 0.177
        & 0.255 & 0.162 & 0.303 & 0.460
        & 0.229 & 0.158 & 0.264 & 0.393
        & 0.230
        \\
\bottomrule
    \end{tabular}}
\caption{Complete LP results on ICEWS21-zero and ICEWS22-zero. We also report PPT and ICL's performance.}
\label{tab: tid LP results first two dataset}
\end{table*}
\begin{table}[htbp]
    \centering
    \resizebox{\columnwidth}{!}{
    \large\begin{tabular}{@{}lcccc cccc c @{}}
\toprule
        \textbf{Datasets} & \multicolumn{9}{c}{\textbf{ACLED-zero}} \\
        & \multicolumn{4}{c}{Zero-Shot Relations} & \multicolumn{4}{c}{Seen Relations} & Overall\\
\cmidrule(lr){2-5} \cmidrule(lr){6-9} \cmidrule(lr){10-10}
        \textbf{Model} & MRR & Hits@1 & Hits@3 & Hits@10 & MRR & Hits@1 & Hits@3 & Hits@10 & MRR\\
\midrule 
        CyGNet 
        & 0.487 & 0.349  & 0.565 & \textbf{0.791}
        & \textbf{0.751}	& 0.663	& \textbf{0.827}	& 0.903
        & 0.717
         \\
        CyGNet+
        & \textbf{0.533} & \textbf{0.418} & \textbf{0.592} & 0.753 
        & \textbf{0.751} & \textbf{0.664} & 0.821 & \textbf{0.906}
        & \textbf{0.723}
        \\
\midrule
        TANGO-T 
        & 0.052 & 0.021 & 0.049	& 0.101
        & 0.774	& 0.701	& 0.826	& 0.900
        & 0.681
        \\
        TANGO-T+ 
        & \textbf{0.525}	& \textbf{0.393} & \textbf{0.606} & \textbf{0.746} 
        & \textbf{0.775}	& \textbf{0.702} & \textbf{0.827} & \textbf{0.901}
        & \textbf{0.743}
        \\
\midrule       
       TANGO-D
       & 0.021		& 0.003 & 0.017		& 0.049
       & \textbf{0.777}	& \textbf{0.701} & \textbf{0.833} & \textbf{0.907} 
       & 0.679
        \\
       TANGO-D+
       & \textbf{0.491}	& \textbf{0.348} & \textbf{0.560}		& \textbf{0.791} 
       & 0.760	& 0.678 & 0.818 & 0.901  
       & \textbf{0.725}
        \\
\midrule
        RE-GCN
        & 0.441	& 0.332 & 0.466	& 0.718
        & 0.730	& \textbf{0.653} & 0.783 & 0.865
        & 0.693
        \\
        RE-GCN+ 
        & \textbf{0.529}	& \textbf{0.393} & \textbf{0.612} & \textbf{0.784} 
        & \textbf{0.731}	& 0.650 & \textbf{0.789} & \textbf{0.876}
        & \textbf{0.705}
        \\
\midrule
        TiRGN
        & 0.478	& 0.330	& 0.572	& 0.745
        & \textbf{0.754}	& 0.678	& 0.806	& \textbf{0.886}
        & 0.718
        \\
        TiRGN+ 
        & \textbf{0.548}	& \textbf{0.436}	& \textbf{0.607}	& \textbf{0.750}	
        & \textbf{0.754}	& \textbf{0.679}	& \textbf{0.807}	& 0.885
        & \textbf{0.727}
        \\
\midrule
RETIA 
        & 0.499	& 0.360	& 0.586	& 0.795	
        & 0.782	& 0.701	& \textbf{0.844}	& 0.924
        & 0.745
        \\
        RETIA+ 
        & \textbf{0.557}	& \textbf{0.408}	& \textbf{0.676}	& \textbf{0.814}	
        & \textbf{0.783}	& \textbf{0.703}	& 0.842	& \textbf{0.925}
        & \textbf{0.754}
        \\
\midrule
        CENET
        & 0.419	& 0.297	& 0.522	& 0.593	
        & 0.753	& 0.682	& 0.808	& 0.869
        & 0.710
        \\
        CENET+ 
        & \textbf{0.591}	& \textbf{0.451}	& \textbf{0.687}	& \textbf{0.844}	
        & \textbf{0.779}	& \textbf{0.692}	& \textbf{0.849}	& \textbf{0.912}
        & \textbf{0.755}
        \\
\midrule
        PPT
        & 0.532 & 0.388 & 0.651 & 0.787
        & 0.782 & 0.693 & 0.842 & 0.942
        & 0.748
        \\
\midrule 
        ICL
        & 0.537 & 0.452 & 0.620 & 0.661
        & 0.736 & 0.668 & 0.794 & 0.853
        & 0.709
        \\
\bottomrule
    \end{tabular}}
\caption{Complete LP results on ACLED-zero. We also report PPT and ICL's performance.}
\label{tab: tid LP results}
\end{table}
\section{Complete Ablation Study Results}
\label{app: ablation}
We report the complete ablation study results in Table \ref{tab: complete ablation}.
\begin{table*}[htbp]
    \centering
    \resizebox{\textwidth}{!}{
    \large\begin{tabular}{@{}lccc ccc c ccc ccc c ccc ccc c@{}}
\toprule
        \textbf{Datasets} & \multicolumn{7}{c}{\textbf{ACLED-zero}} & \multicolumn{7}{c}{\textbf{ICEWS21-zero}} & \multicolumn{7}{c}{\textbf{ICEWS22-zero}} \\
        & \multicolumn{3}{c}{Zero-Shot Relations} & \multicolumn{3}{c}{Seen Relations} & Overall & \multicolumn{3}{c}{Zero-Shot Relations} & \multicolumn{3}{c}{Seen Relations} & Overall &\multicolumn{3}{c}{Zero-Shot Relations} & \multicolumn{3}{c}{Seen Relations} & Overall \\
\cmidrule(lr){2-4} \cmidrule(lr){5-7} \cmidrule(lr){8-8} \cmidrule(lr){9-11} \cmidrule(lr){12-14} \cmidrule(lr){15-15}\cmidrule(lr){16-18} \cmidrule(lr){19-21} \cmidrule(lr){22-22}
        \textbf{Model} & MRR & Hits@1 & Hits@10 & MRR & Hits@1 & Hits@10 & MRR & MRR & Hits@1 & Hits@10 & MRR & Hits@1 & Hits@10 & MRR & MRR & Hits@1 & Hits@10 & MRR & Hits@1 & Hits@10 & MRR\\
\midrule 
        CyGNet+ 
        & \textbf{0.533} & \textbf{0.418} & \textbf{0.753}
        & 0.751  & \textbf{0.664} & \textbf{0.906}
        & \textbf{0.723} 
        & \textbf{0.201} &	\textbf{0.103} & \textbf{0.415}
        & \textbf{0.258} & \textbf{0.162} & \textbf{0.447}
        & \textbf{0.257}
        & \textbf{0.286} & \textbf{0.167} & \textbf{0.542}
        & \textbf{0.315} & 0.200 & 0.545
        & \textbf{0.314}
         \\
        \ - ERD
        & 0.502	& 0.386 & 0.743
        & 0.748 & 0.660 & 0.902
        & 0.716
        & 0.198 & 0.102 & 0.379	
        & 0.252	& 0.161 & 0.429
        & 0.251
        & 0.250	& 0.136 & 0.503
        & 0.314	& 0.198 & \textbf{0.546}
        & 0.311
        \\
        \ - RHL
        & 0.503		& 0.356 & 0.751
        & \textbf{0.752} & 0.663 & 0.901
        & 0.720
        & 0.199	& 0.100 & 0.398
        & 0.256	& 0.159 & 0.445
        & 0.255
        & 0.268	& 0.144 & 0.536
        & 0.297	& 0.181 & 0.531
        & 0.296
        \\
        \ T5-3B
        & 0.511	& 0.414 & 0.684
        & \textbf{0.752}	& 0.663 & 0.905
        & 0.721
        & 0.117	& 0.068 & 0.186
        & 0.204	& 0.127 & 0.348
        & 0.202
        & 0.257	& 0.135 & 0.521
        & \textbf{0.315} & \textbf{0.201} & 0.540
        & 0.313
        \\
\midrule
        TANGO-T+
        & 0.525	& 0.393 & 0.764
        & \textbf{0.775}	& 0.702 & \textbf{0.901}
        & \textbf{0.743}
        & \textbf{0.216} & \textbf{0.125} & 0.395	
        & \textbf{0.280} & 0.186 & 0.466
        & \textbf{0.279}
        & \textbf{0.326} & \textbf{0.198} & \textbf{0.578}	
        & \textbf{0.363} & \textbf{0.251} & \textbf{0.585}
        & \textbf{0.362}
        \\
        \ - ERD
        & 0.533		& 0.408    & \textbf{0.770}
        & 0.772 & 0.692 & 0.898
        & 0.741
        & 0.214	& 0.122 & 0.389
        & \textbf{0.280}	& \textbf{0.187} & 0.465	
        & \textbf{0.279}
        & 0.320	 & 0.193 & 0.576
        & 0.362	 & 0.250 & 0.584
        & 0.360
        \\
        \ - RHL
        & 0.506	& 0.374 & 0.749
        & 0.755	& \textbf{0.704} & \textbf{0.901}
        & 0.740
        & 0.213	& 0.118 & \textbf{0.407}
        & 0.277	& 0.181 & \textbf{0.469}
        & 0.276
        & 0.309	& 0.190 & 0.574
        & \textbf{0.363}	& 0.250 & 0.584
        & 0.361
        \\
        \ T5-3B
        & \textbf{0.544}	& \textbf{0.425} & 0.769	
        & 0.771 & 0.697 & 0.896
        & 0.742
        & 0.206 & 0.119 & 0.375
        & 0.274	& 0.182 & 0.454
        & 0.273
        & 0.323 & 0.193 & 0.576
        & 0.359 & 0.246 & 0.579
        & 0.358
        \\
\midrule       
       TANGO-D+
        & \textbf{0.491}	& 0.348 & \textbf{0.791}
        & \textbf{0.760}	& \textbf{0.678} & \textbf{0.901}
        & \textbf{0.725}
        & \textbf{0.212} & \textbf{0.122} & \textbf{0.400}		
        & \textbf{0.268} & \textbf{0.175} & \textbf{0.453}
        & \textbf{0.267}
        & \textbf{0.311} & \textbf{0.186} & 0.574
        & \textbf{0.350} & \textbf{0.239} & \textbf{0.570}
        & \textbf{0.348}
        \\
       \ - ERD
        & \textbf{0.491} & \textbf{0.350} & 0.771
        & 0.702	& 0.578 & 0.898
        & 0.675
        & 0.205	& 0.111 & 0.398
        & 0.267	& 0.174 & 0.449
        & 0.266
        & 0.285	& 0.159 & 0.541
        & 0.328	& 0.213 & 0.550
        & 0.326
        \\
        \ - RHL
        & 0.490 & 0.344 & 0.772
        & 0.725	& 0.628 & 0.890
        & 0.695
        & 0.197	& 0.107 & 0.390
        & 0.224	& 0.132 & 0.412
        & 0.224
        & 0.296	& 0.175 & 0.552	
        & 0.324	& 0.212 & 0.547
        & 0.323
        \\
        \ T5-3B
        & 0.490		& 0.341 & 0.786
        & 0.701	& 0.576 & 0.897
        & 0.674
        & 0.204	& 0.109 & 0.393
        & 0.223	& 0.131 & 0.408
        & 0.222
        & 0.308	& 0.177 & \textbf{0.582}
        & 0.284 & 0.173 & 0.510
        & 0.285
        \\
\midrule
        RE-GCN+
        & \textbf{0.529}	& 0.393 & \textbf{0.784}	
        & \textbf{0.731}	& \textbf{0.650} & \textbf{0.876}
        & \textbf{0.705}
        & \textbf{0.214} & 0.117 & \textbf{0.406}		
        & \textbf{0.280} & \textbf{0.188} & \textbf{0.456}
        & \textbf{0.279}
        & \textbf{0.324} & \textbf{0.194} & \textbf{0.595}
        & \textbf{0.357} & \textbf{0.244} & \textbf{0.573}
        & \textbf{0.356}
        \\
        \ - ERD
        & 0.489	& 0.375 & 0.724
        & 0.730	& \textbf{0.650} & 0.865
        & 0.699
        & 0.211	& 0.119 & 0.397
        & 0.277	& 0.185 & 0.454
        & 0.276
        & 0.294	& 0.168 & 0.560
        & 0.354	& 0.242 & 0.571
        & 0.352
        \\
        \ - RHL
        & 0.519	& \textbf{0.396} & 0.757
        & 0.726 & 0.646 & 0.836
        & 0.699
        & 0.213	& 0.119 & 0.405
        & 0.277	& 0.185 & 0.455
        & 0.276
        & 0.317	& 0.184 & 0.589
        & 0.350	& 0.241 & 0.562
        & 0.349
        \\
        \ T5-3B
        & 0.504 & 0.361 & 0.767
        & 0.721 & 0.638 & 0.864
        & 0.693
        & 0.211	& \textbf{0.121} & 0.384
        & 0.259	& 0.171 & 0.427
        & 0.258
        & 0.301 & 0.174 & 0.577
        & 0.354 & 0.243 & 0.570
        & 0.352
        \\
\midrule
        TiRGN+
        & \textbf{0.548} & \textbf{0.436} & 0.750
        & \textbf{0.754} & \textbf{0.679} & 0.885
        & \textbf{0.727}
        & \textbf{0.221} & \textbf{0.130} & \textbf{0.410}	
        & \textbf{0.279} & \textbf{0.185} & \textbf{0.463}
        & \textbf{0.278}
        & \textbf{0.333} & \textbf{0.203} & \textbf{0.602}
        & \textbf{0.353} & \textbf{0.240} & \textbf{0.577}
        & \textbf{0.352}
        \\
        \ - ERD
        & 0.480	& 0.387 & 0.673
        & 0.747 & 0.669 & 0.882
        & 0.713
        & 0.211	& 0.120 & 0.387	
        & 0.275	& 0.181 & 0.460 
        & 0.274
        & 0.282	& 0.157 & 0.544
        & \textbf{0.353}	& \textbf{0.240} & 0.576
        & 0.350
        \\
        \ - RHL
        & 0.515		& 0.400 & \textbf{0.753}
        & 0.752 & 0.675 & \textbf{0.887}
        & 0.721
        & 0.215	& 0.124 & 0.391	
        & 0.277	& 0.183 & 0.461
        & 0.276
        & 0.320	& 0.190 & 0.593
        & 0.350	& 0.239 & 0.569
        & 0.349
        \\
        \ T5-3B
        & 0.498 & 0.389 & 0.722
        & 0.749 & 0.675 & 0.879
        & 0.717
        & 0.208	& 0.118 & 0.392
        & 0.271 & 0.180 & 0.448
        & 0.270
        & 0.325 & 0.189 & 0.594
        & 0.345 & 0.233 & 0.565
        & 0.344
        \\
\midrule
        RETIA+
        & \textbf{0.557}	& \textbf{0.408} & \textbf{0.814}	
        & \textbf{0.783}    & \textbf{0.703} & \textbf{0.925}
        & \textbf{0.754}
        &&
                \multicolumn{5}{c}{\multirow{4}{*}{\text{>> 120 Hours Timeout}}}
        &
        & \textbf{0.331}	& \textbf{0.201} & \textbf{0.597}
        & \textbf{0.358}	& \textbf{0.247} & 0.578
        & \textbf{0.357}
        \\
        \ - ERD
        & 0.519	& 0.391 & 0.765
        & 0.777	& 0.692 & 0.917
        & 0.744
        &&&&&&
        &
        & 0.292 & 0.163 & 0.562
        & 0.354	& 0.242 & 0.576
        & 0.352
        \\
        \ - RHL
        & 0.529	& 0.368 & 0.796
        & 0.782	& 0.701 & 0.923
        & 0.749
        &&&&&&
        &
        & 0.318	& 0.191 & 0.583
        & 0.357	& 0.244 & \textbf{0.580}
        & 0.355
        \\
        \ T5-3B
        & 0.512	& 0.385 & 0.766
        & 0.776 & 0.690 & 0.917
        & 0.742
        &&&&&&
        &
        & 0.330 & 0.200 & 0.595
        & 0.353	& 0.242 & 0.573
        & 0.352
        \\
\midrule
        CENET+
        & \textbf{0.591}	& \textbf{0.451} & \textbf{0.844}	
        & \textbf{0.779}	& \textbf{0.692} & \textbf{0.912}
        & \textbf{0.755}
        & \textbf{0.335}	& \textbf{0.162} & 0.659
        & \textbf{0.396}	& \textbf{0.239} & \textbf{0.688}
        & \textbf{0.395}
        & \textbf{0.564}	& \textbf{0.432} & \textbf{0.801}
        & \textbf{0.571}	& \textbf{0.451} & 0.773
        & \textbf{0.570}
        \\
        \ - ERD
        & 0.526		& 0.373 & 0.785
        & 0.737	    & 0.653 & 0.870
        & 0.710
        & 0.321	& 0.156 & \textbf{0.665}
        & 0.374	& 0.216 & 0.683
        & 0.373
        & 0.542 & 0.388 & 0.799
        & 0.570	& 0.448 & \textbf{0.774}
        & 0.568
        \\
        \ - RHL
        & 0.445	 & 0.367 & 0.565
        & 0.754	 & 0.685 & 0.862
        & 0.714
        & 0.232	& 0.128 & 0.446
        & 0.290	& 0.202 & 0.469
        & 0.289
        & 0.295	& 0.168 & 0.560
        & 0.370	& 0.262 & 0.588
        & 0.367
        \\
        \ T5-3B
        & 0.568	& 0.426 & 0.819
        & 0.736 & 0.646 & 0.900
        & 0.714
        & 0.303	& 0.158 & 0.568
        & 0.330	& 0.203 & 0.712
        & 0.329
        & 0.550	& 0.413 & 0.798
        & 0.555 & 0.431 & 0.765
        & 0.554
        \\
\bottomrule
    \end{tabular}}
\caption{Complete results of ablation studies.}
\label{tab: complete ablation}
\end{table*}
\section{Complete Results of Previous LM-Enhanced TKGF Model}
\label{app: previous LM}
We report the complete results of previous LM-enhanced TKGF models in Table \ref{tab: tid LP results first two dataset} and \ref{tab: tid LP results}.
\section{Further Discussion about RHL}
\label{app: rhl}
In RHL, temporal relation patterns are captured by only using LLM-based relation representations. Since for all relations (whether zero-shot or not), their LLM-based representations contain semantic information extracted from the same LLM, the learned HPN can do reasonable relation history prediction even with an input of unseen zero-shot relation. If we learn hidden representations for each relation based on graph contexts (as most TKGF models do), zero-shot relations cannot be easily processed by HPN anymore. In this case, zero-shot relations will not have a meaningful representation without any observed associated fact, and therefore, HPN cannot detect its meaning and will fail to find reasonable relation history.
\section{Failure Case Discussion}
From Table \ref{tab: ablation}, we observe several failure cases when the complete zrLLM is implemented, e.g., (1) TANGO-T+ without ERDs show a slightly better zero-shot result on ACLED-zero compared with the complete TANGO-T+; (2) TANGO-T+ does not witness an improvement over the seen relations on ICEWS21-zero compared with TANGO-T+ without RHL. We attribute such failure cases to the characteristics of the considered TKGF models. As highlighted in Sec. \ref{sec: exp setup}, our goal is to use zrLLM to enhance TKGF model performance over zero-shot relations while maintaining strong performance over seen relations. By carefully comparing the overall performance of zrLLM-enhanced models with their ablated variants, e.g., -ERD, we find that the complete version of zrLLM with ERDs, RHL and T5-11B can always achieve the best overall performance, which aligns to our motivation. The small number of failure cases caused by several baseline TKGF methods cannot overturn the merit brought by the modules of zrLLM.
\section{Related Work Details}
\label{app: related work}
\paragraph{Traditional TKG Forecasting Methods.}
As discussed in Sec. \ref{sec: intro}, traditional TKGF methods are trained to forecast the facts containing the KG relations (and entities) seen in the training data, regardless of the case where zero-shot relations (or entities) appear as new knowledge arrives\footnote{Some works of traditional TKGF methods, e.g., TANGO \cite{DBLP:conf/emnlp/HanDMGT21}, have discussions about models' ability to reason over the facts regarding unseen entities. Note that this is not their main focus but an additional demonstration to show their models' inductive power, i.e., these models are not designed for inductive learning on TKGs.}. These methods can be categorized into two types: embedding-based and rule-based. Embedding-based methods learn hidden representations of KG relations and entities (some also learn time representations), and perform link forecasting by inputting learned representations into a score function for computing scores of fact quadruples. Most existing embedding-based methods, e.g., \cite{DBLP:conf/emnlp/JinQJR20,DBLP:conf/emnlp/HanDMGT21,DBLP:conf/sigir/LiJLGGSWC21,DBLP:conf/ijcai/LiS022,DBLP:conf/icde/Liu0X0023}, learn evolutional entity and relation representations by jointly employing graph neural networks \cite{DBLP:conf/iclr/KipfW17} and recurrent neural structures, e.g., GRU \cite{DBLP:conf/emnlp/ChoMGBBSB14}. Historical TKG information are recurrently encoded by the models to produce the temporal sequence-aware evolutional representations for future prediction. Some other approaches \cite{DBLP:conf/iclr/HanCMT21,sun-etal-2021-timetraveler,DBLP:conf/acl/LiJGLGWC20} start from each LP query and traverse the temporal history in a TKG to search for the prediction answer. Apart from them, CyGNet \cite{DBLP:conf/aaai/ZhuCFCZ21} achieves forecasting purely based on the appearance of historical facts. 
Another recent work CENET \cite{DBLP:conf/aaai/XuO0F23} trains contrastive representations of LP queries to identify
highly correlated entities in either historical or non-historical facts. 
Compared with the rapid advancement in developing embedding-based TKGF methods, rule-based TKGF has still not been extensively explored. One popular rule-based TKGF method is TLogic \cite{DBLP:conf/aaai/LiuMHJT22}. It extracts temporal logic rules from TKGs and uses a symbolic reasoning module for LP. Based on it, ALRE-IR \cite{DBLP:conf/emnlp/MeiYCJ22} proposes an adaptive logical rule embedding model to encode temporal logical rules into rule representations. This makes ALRE-IR both a rule-based and an embedding-based method. Experiments in TLogic and ALRE-IR have proven that rule-based TKGF methods have strong ability in reasoning over zero-shot unseen entities connected by the seen relations, however, they are not able to handle unseen relations since the learned rules are strongly bounded by the observed relations. In our work, we implement zrLLM on embedding-based TKGF models because (1) embedding-based methods are much more popular; (2) zrLLM utilizes LLM to generate relation representations, which is more compatible with embedding-based methods.

\paragraph{Inductive Learning on TKGs.}
Inductive learning on TKGs has gained increasing interest. It refers to developing models that can handle the relations and entities unseen in the training data. TKG inductive learning methods can be categorized into two types. The first type of works focuses on reasoning over unseen entities \cite{ding2022few,DBLP:conf/nips/0004LSLLYA22,DBLP:conf/pkdd/DingWLMT23,DBLP:conf/sigir/ChenXS0D23}, while the second type of methods aims to deal with the unseen relations \cite{DBLP:conf/akbc/MirtaheriR0MG21,DBLP:conf/ijcnn/DingHWMHT23,DBLP:journals/datamine/MaMMZL023}. 
Most of inductive learning methods are based on few-shot learning (e.g., FILT \cite{ding2022few}, MetaTKGR \cite{DBLP:conf/nips/0007TYL19}, FITCARL \cite{DBLP:conf/pkdd/DingWLMT23}, OAT \cite{DBLP:conf/akbc/MirtaheriR0MG21}, MOST \cite{DBLP:conf/ijcnn/DingHWMHT23} and OSLT \cite{DBLP:journals/datamine/MaMMZL023}). They first compute inductive representations of newly-emerged entities or relations based on $K$-associated facts ($K$ is a small number, e.g., 1 or 3) observed during inference, and then use them to predict the facts regarding few-shot elements. One limitation of these works is that the inductive representations cannot be learned without the $K$-shot examples, making them hard to solve the zero-shot problems. Different from few-shot learning methods, SST-BERT \cite{DBLP:conf/sigir/ChenXS0D23} pre-trains a time-enhanced BERT \cite{DBLP:conf/naacl/DevlinCLT19} for TKG reasoning. It achieves inductive learning over unseen entities but has not shown its ability in reasoning zero-shot relations. Another recent work MTKGE \cite{DBLP:conf/www/ChenXS0D23} is able to concurrently deal with both unseen entities and relations. However, it requires a support graph containing a substantial number of data examples related to the unseen entities and relations, which is far from the zero-shot problem that we focus on.

\paragraph{TKG Reasoning with Language Models.}
\label{app: TKG with LLM}
Recently, more and more works have introduced LMs into TKG reasoning. SST-BERT \cite{DBLP:conf/sigir/ChenXS0D23} generates a small-scale pre-training corpus based on the training TKGs and pre-trains an LM for encoding TKG facts. The encoded facts are then fed into a scoring module for LP. ECOLA \cite{han-etal-2023-ecola} aligns facts with additional fact-related texts and proposes a joint training framework that enhances TKG reasoning with BERT-encoded language representations. 
PPT \cite{DBLP:conf/acl/XuLPJP23} converts TKGF into the pre-trained LM masked token prediction task and finetunes a BERT for TKGF. It directly input TKG facts into the LM for answer prediction.
Apart from them, one recent work \cite{DBLP:journals/corr/abs-2305-10613} explores the possibility of using in-context learning (ICL) \cite{DBLP:conf/nips/BrownMRSKDNSSAA20} with LLMs to make predictions about future facts without fintuning. 
Another recent work GenTKG \cite{DBLP:journals/corr/abs-2310-07793} finetunes an LLM, i.e., Llama2-7B \cite{DBLP:journals/corr/abs-2302-13971}, and let the LLM directly generate the LP answer in TKGF. It mines temporal logical rules and uses them to retrieve historical facts for prompt generation. 

Although the above-mentioned works have shown success of LMs in TKG reasoning, they have limitations: (1) None of these works has studied whether LMs can be used to better reason the zero-shot relations. (2) By only using ICL, LLMs are beaten by traditional TKG reasoning methods in performance \cite{DBLP:journals/corr/abs-2305-10613}. The performance can be greatly improved by finetuning LLMs (as in GenTKG \cite{DBLP:journals/corr/abs-2310-07793}), but finetuning LLMs requires huge computational resources. (3) Since LMs, e.g., BERT and Llama2, are pre-trained with a huge corpus originating from diverse information sources, it is inevitable that they have already seen the world knowledge before they are used to solve TKG reasoning tasks. Most popular TKGF benchmarks are extracted from the TKGs constructed before 2020, e.g., ICEWS14, ICEWS18 and ICEWS05-15 \cite{DBLP:conf/emnlp/JinQJR20}. The facts inside are based on the world knowledge before 2019, which means LMs might have encountered them in their training corpus, posing a threat of information leak to the LM-driven TKG reasoning models. To this end, we (1) draw attention to studying the impact of LMs on zero-shot relational learning in TKGs; (2) make a compromise between performance and computational efficiency by not fintuning LMs or LLMs but adapting the LLM-provided semantic information to non-LM-based TKGF methods; (3) construct new benchmarks where the facts are all happening from 2021 to 2023, which avoids the possibility of information leak when we utilize T5-11B that was released in 2020.

\end{document}